\documentclass[preprint]{imsart}

\RequirePackage{amsthm,amsmath,amsfonts,amssymb}
\RequirePackage[numbers,sort&compress]{natbib}
\RequirePackage[colorlinks,citecolor=blue,urlcolor=blue]{hyperref}
\RequirePackage{graphicx}
\RequirePackage{mathtools}
\RequirePackage{dsfont}
\RequirePackage{bm}
\RequirePackage{enumitem}
\RequirePackage{color,soul}

\pubyear{2025}
\arxiv{2501.01963}
\volume{TBA}
\issue{TBA}
\firstpage{1}
\lastpage{28}

\startlocaldefs
\theoremstyle{plain}
\newtheorem{axiom}{Axiom}
\newtheorem{claim}[axiom]{Claim}
\newtheorem{theorem}{Theorem}[section]
\newtheorem{proposition}{Proposition}[section]

\theoremstyle{remark}
\newtheorem{remark}{Remark}
\newtheorem{definition}[theorem]{Definition}
\newtheorem{example}{Example}

\newcommand{\defeq}{\vcentcolon=}

\DeclarePairedDelimiterX{\infdivx}[2]{(}{)}{%
  #1\;\delimsize\|\;#2%
}
\newcommand{\infdiv}{D\infdivx} %\infdiv{P}{Q}

\makeatletter
\newcommand{\@giventhatstar}[2]{\left(#1\;\middle|\;#2\right)}
\newcommand{\@giventhatnostar}[3][]{#1(#2\;#1|\;#3#1)}
\newcommand{\giventhat}{\@ifstar\@giventhatstar\@giventhatnostar}
\makeatother
\newcommand\restr[2]{{
  \left.\kern-\nulldelimiterspace 
  #1 
  \littletaller 
  \right|_{#2} 
  }}
\newcommand{\littletaller}{\mathchoice{\vphantom{\big|}}{}{}{}}

\def\D{\mathrm{d}}
\def\s{\sigma}

\def\AA{\mathsf{A}}
\def\BB{\mathsf{B}}
\def\CC{\mathsf{C}}
\def\DD{\mathsf{D}}
\def\EE{\mathsf{E}}

\def\TT{\mathsf{T}}

\def\A{\mathcal{A}}

\def\F{\mathcal{F}}
\def\G{\mathcal{G}}

\def\I{\mathcal{I}}
\def\J{\mathcal{J}}
\def\L{\mathcal{L}}
\def\S{\mathcal{S}}
\def\T{\mathcal{T}}

\def\PP{\mathcal{P}}
\def\QQQ{\mathcal{Q}}

\def\S{\mathcal{S}}
\def\X{\mathcal{X}}

\def\R{\mathbb{R}}

\def\Var{\mathrm{Var}}
\def\Cov{\mathrm{Cov}}

\def\beq{\begin{equation}}
\def\eeq{\end{equation}}
\def\lb{\label}

\def\bde{\boldsymbol\delta}
\def\bla{\boldsymbol\lambda}
\def\bmu{\boldsymbol\mu}
\def\bpi{\boldsymbol\pi}
\def\bSi{\boldsymbol\Sigma}
\def\tbla{\tilde{\bla}}
\def\hbla{\hat{\bla}}

\def\hmu{\hat{\mu}}
\def\hbmu{\hat{\bmu}}
\def\tA{\tilde{\A}}
\def\tP{\tilde{\P}}

\def\J{\mathbf J}
\def\f{\mathbf f}
\def\H{\mathbf H}

\def\E{{\mathbf E}}

\def\P{{\mathbf P}}
\def\QQ{{\mathbf Q}}

\def\W{{\mathbf W}}

\def\1{\mathds{1}} %\usepackage{dsfonts}

\endlocaldefs

\begin{document}

\begin{frontmatter}
\title{Statistical learning does not always entail knowledge}
%\title{A sample article title with some additional note\thanksref{t1}}
\runtitle{Statistical learning does not always entail knowledge}
%\thankstext{T1}{A sample additional note to the title.}

\begin{aug}
%%%%%%%%%%%%%%%%%%%%%%%%%%%%%%%%%%%%%%%%%%%%%%%
%% Only one address is permitted per author. %%
%% Only division, organization and e-mail is %%
%% included in the address.                  %%
%% Additional information can be included in %%
\author[A]{\fnms{Daniel Andrés}~\snm{Díaz-Pachón}\ead[label=e1]{ddiaz3@miami.edu}\orcid{0000-0001-6281-1720}},
\author[A]{\fnms{H. Renata}~\snm{Gallegos}\ead[label=e2]{h.gallegos@med.miami.edu}},
\author[B]{\fnms{Ola}~\snm{H\"ossjer}\ead[label=e3]{ola@math.su.se}\orcid{0000-0003-2767-8818}},
\and
\author[C]{\fnms{J. Sunil}~\snm{Rao}\ead[label=e4]{js-rao@umn.edu}\orcid{0000-0002-6450-3200}},
%%%%%%%%%%%%%%%%%%%%%%%%%%%%%%%%%%%%%%%%%%%%%%
%% Addresses                                %%
%%%%%%%%%%%%%%%%%%%%%%%%%%%%%%%%%%%%%%%%%%%%%%
\address[A]{Division of Biostatistics, University of Miami\printead[presep={ ,\ }]{e1,e2}}
\address[B]{Department of Mathematics, Stockholm University\printead[presep={,\ }]{e3}}
\address[C]{Department of Biostatistics, University of Minnesota\printead[presep={,\ }]{e4}}
\end{aug}

\begin{abstract}
In this paper, we study learning and knowledge acquisition (LKA) of an agent about a proposition that is either true or false. We use a Bayesian approach, where the agent receives data to update his beliefs about the proposition according to a posterior distribution. The LKA is formulated in terms of active information, with data representing external or exogenous information that modifies the agent's beliefs. It is assumed that data provide details about a number of features that are relevant to the proposition. We show that this leads to a Gibbs distribution posterior, which is in maximum entropy relative to the prior, conditioned on the side constraints that the data provide in terms of the features. We demonstrate that full learning is sometimes not possible and full knowledge acquisition (KA) is never possible when the number of extracted features is too small. We also distinguish between primary learning (receiving data about features of relevance for the proposition) and secondary learning (receiving data about the learning of another agent). We argue that this type of secondary learning does not represent true KA. Our results have implications for statistical learning algorithms, and we claim that such algorithms do not always generate true knowledge. The theory is illustrated with several examples.
\end{abstract}



\end{frontmatter}
%%%%%%%%%%%%%%%%%%%%%%%%%%%%%%%%%%%%%%%%%%%%%%
%% Please use \tableofcontents for articles %%
%% with 50 pages and more                   %%
%%%%%%%%%%%%%%%%%%%%%%%%%%%%%%%%%%%%%%%%%%%%%%
%\tableofcontents

\section{Introduction}\label{Sec:Intro}

\subsection{Learning and knowledge acquisition}\lb{Sec:LKAIntro}

In the current era of scientific computing, when large language models have seemingly achieved surprising levels of understanding and discussions about artificial general intelligence are as abundant as nebulous, proper definitions that can be accurately quantified are conspicuous by their absence. For instance, what do we mean by ``understanding'' and ``intelligence'' in the previous paragraph? If explainable AI is going to explain anything, it does require clear concepts capable of guiding the discussion to reach valid conclusions. Philosophers usually define knowledge as ``justified true belief'' \cite{Gettier1963, IchikawaSteup2018, Schwitzgebel2021}. This means that an agent $\A$ \textit{knows} a proposition $p$ if the following three properties are satisfied:
\begin{enumerate}[label = {\bf LK\arabic*}, nosep]
	\item\label{K1} $\A$ believes $p$,
	\item\label{K2} $p$ is true,
	\item\label{K3} $\A$'s belief about $p$ is justified.
\end{enumerate}
If only properties \ref{K1} and \ref{K2} are satisfied, $\A$ \textit{learns} $p$. Clearly, acquiring knowledge requires more than learning. Therefore, even before further theoretical developments, we obtain a simple but revealing fact: 
\begin{claim}\lb{Claim}
	Statistical learning does not always entail knowledge. 
\end{claim}
A mathematical phrasing of Claim 1 is that even when statistical learning generates true beliefs (\ref{K1}-\ref{K2}), these beliefs are not necessarily justified (\ref{K3}). The mathematical formulation of learning and knowledge acquisition (LKA), based on \ref{K1}-\ref{K3}, was introduced in \cite{HossjerDiazRao2022}. The main idea is that agent $\A$ (for instance a large language algorithm) uses data $\DD$ to learn and acquire knowledge about $p$. This is described with a mixed Bayesian-frequentist model, where beliefs in \ref{K1} correspond to a posterior distribution, whereas frequentist concepts are needed to formalize \ref{K2}-\ref{K3}. This approach has already been applied to determine which cases of cosmological fine-tuning can be known \cite{DiazHossjerMathew2024} (see also \cite{DiazHossjerMarks2021, DiazHossjerMarks2023}). Our approach to LKA goes further in four ways: 

\begin{description}[nosep]
    \item[(i)] We develop the notion of discernment introduced in \cite{HossjerDiazRao2022}, further quantifying how it imposes limits on LKA. Mathematically, discernment corresponds to a $\s$-field that describes how well different possible explanations of $p$ can be separated. This $\s$-field sets limits on the posterior distribution (the beliefs of $\A$). In this article, we give very explicit conditions when the $\s$-field is too coarse to warrant full learning and full knowledge acquisition (KA), respectively.  
    \item[(ii)] We focus on learning through feature extraction and Gibbs distributions. This is a natural and powerful approach, which mathematically corresponds to a method of moments and quasi-Bayesian approach. That is, the posterior distribution is not obtained directly through Bayes Theorem. Instead, data $\DD$ is used to update prior beliefs by matching expected and observed features.  
    \item[(iii)] We motivate Claim \ref{Claim} through multiple examples and results in which knowledge cannot be acquired, even if {\it full} learning is attained. Although this was pointed out in \cite{HossjerDiazRao2022}, in this article we make use of (ii) and give examples where observed features are enough for full learning, whereas unobserved, hidden features would be required in order to acquire full knowledge. Hypothetically, one may imagine that these hidden features are those that require intuition and creative thinking. The fact that some features are hidden typically implies that the possible explanations of $p$ are unidentifiable, making full KA impossible. This highlights that the quasi-Bayesian Gibbs distribution approach, outlined in (ii), is more than a technical extension of traditional Bayesian inference. It is rather a very useful tool for LKA analyses, since it helps to quantify the limits of knowledge acquisition.    
    \item[(iv)] We introduce the concepts of primary and secondary learning. These concepts are applicable both for traditional Bayesian modeling and for quasi-Bayesian modeling with Gibbs distributions. Primary learning is based on processing data $\DD$ about proposition $p$ in order to form beliefs, whereas secondary learning uses secondary data $\tilde{\DD}$ in the sense of learning what other agents learn about $p$. In between %these two types of learning 
    is synthetic primary learning, where artificially generated data $\DD^\prime$ of relevance for $p$ are used in order to form beliefs about $p$. We will argue that synthetic primary learning as well as secondary learning, may be subject to bias. Since a lot of statistical learning is based on indirect data sources, this is also another motivation for Claim 1.        
\end{description}

\subsection{Active information}
To describe (i)-(iv) in more detail, we introduce local measures of information. Despite Shannon's information theory almost exclusive focus on global averages such as entropy, mutual information, relative entropy, etc., recent decades have seen a resurgence of unaveraged measures of information like local active information storage and local transfer entropy. These measures have been used in origin of life \citep{Davies2019, WalkerDavies2013, WibralLizierPriesemann2015}, neuroscience \cite{WibralLizierPriesemann2014, WibralEtAl2014} as well as cancer research and cell communication \cite{McMillenWalkerLevin2022, MooreWalkerLevin2017}. All such measures can be seen as extensions of the more basic active information (AIN), which was originally proposed to measure the amount of exogenous information infused by a programmer in a search, compared to the endogenous information generated by a blind search \citep{DembskiMarks2009b, DembskiMarks2010}. Formally, if the distributions of the outcome of the programmer $\A$ and the blind search $\I$ are represented by two probability measures $\P$ and $\P_0$ defined on the same measurable space $(\X, \F)$, AIN for a specific target $\TT\subset\X$ is defined as 
\begin{align}\label{AIN}
	I^+(\TT) = I^+(\TT;\P_0,\P) = \log \P(\TT) -\log \P_0(\TT),
\end{align}
where we assume 0/0 = 0 by continuity. In particular, if the programmer reaches the target with certainty ($\P(\TT)=1$), then \eqref{AIN} reduces to the self-information of $\TT$. To this point, AIN has been used in several areas. For instance, in genetics, to quantify functional information in genetic sequence data \cite{ThorvaldsenHossjer2023, ThorvaldsenHossjer2024}, and to compare selectively non-neutral models to neutral ones in population genetics, where $\TT$ was the event that a given allele gets fixed \cite{DiazMarks2020b}; in bump-hunting, using machine learning algorithms to find a bump $\TT$ \cite{DiazEtAl2019, LiuEtAl2023}; and in decision theory, to construct hypothesis tests that quantify the amount of information added, or needed, to produce an event $\TT$ \cite{DiazHossjer2022, DiazSaenzRao2020}. 

\subsection{A mixed frequentist-Bayesian framework for LKA}\lb{Sec:AILKA}

Following \cite{HossjerDiazRao2022}, in this article we apply AIN to formalize the concepts \ref{K1}-\ref{K3} behind LKA. To this end, it is assumed that $\X$ is a set of parameters (also referred to as the set of possible worlds) of a statistical model, and we take a mixed Bayesian-frequentist approach. On the one hand, it is postulated that one element $x_0\in\X$ is the true para\-meter value or the true world (a frequentist assumption). On the other hand, uncertainty about $x_0$ is formulated as a probability measure on $\X$ that varies between persons (a Bayesian assumption). More specifically, $\P$ and $\P_0$ represent degrees of beliefs about $x_0\in\X$, of an agent $\A$ and an ignorant person $\I$, respectively. It is assumed that $\A$ acquired data $\DD$ that $\I$ lacks, so that $\P$ and $\P_0$ are posterior and prior distributions on $\X$ that represent degrees of beliefs of $\A$ about $x_0$, after and before he received data. In particular, if we choose $\TT$ as the set of parameter values for which a given proposition $p$ is true, the objective of $\A$ is to use data to learn whether the proposition is true ($x_0\in \TT$) or not ($x_0 \notin \TT$), as quantified by the AIN $I^+(\TT)$ in \eqref{AIN}. In this case, data represent the exogenous information that helps $\A$ to modify his beliefs \ref{K1} about $\TT$ compared to the ignorant person $\I$. KA goes beyond learning since it additionally requires \ref{K3}, that $\A$ learns about the proposition for the right reason. This corresponds to increasingly correct beliefs about $x_0$, not only increasingly correct beliefs of whether $x_0\in\TT$ or not (as for learning). Our approach proposes a very sensible solution to the old dispute between Bayesians and frequentists. We consider propositions and states of reality that are objectively true or false, but LKA are naturally Bayesian. Thus, frequentism accounts for ontology, whereas epistemology is Bayesian. Our definitions differentiate between them; an essential aspect of our theory. Other examples in which ontology is incorporated within a Bayesian framework is when a Bayesian approach is used to test the goodness-of-fit of a model \cite{GelmanShalizi2013} and in Bayesian asymptotic theory, where one parameter value is regarded as the true one \cite{GhosalvanderVaart2017}. However, to the best of our knowledge, a systematic frequentist-Bayesian theory of LKA has not been developed before the work of \cite{HossjerDiazRao2022}. Other approaches to knowledge acquisition appear, for instance, in \cite{Hopkins2002, StoicaStrack2017, Taylor2002}.    

\subsection{The novelties of this article}

Given the framework outlined in Section \ref{Sec:AILKA}, the novelties (i)-(iv) in Section \ref{Sec:LKAIntro} can be phrased as follows. Starting with (i), discernment is a crucial aspect of agent $\A$'s LKA process, which quantifies his ability to separate elements of $\X$ from each other. $\A$'s discernment is typically restricted by the quality of the data he receives, but it is still larger than the ignorant person $\I$'s ability to discern. That is, $\A$'s beliefs $\P$ are measurable on a finer $\sigma$-field of $\X$ than $\I$'s beliefs $\P_0$. We prove general results on how $\A$'s $\sigma$-field affects his potential to learn and acquire knowledge. As for (ii), we assume that data provide $\A$ with details about (modifies his beliefs in) the values of a number of features of relevance for learning proposition $p$. Then $\A$ forms his likelihood in such a way that $\P$ maximizes entropy relative to $\P_0$, among all probability measures on $\X$ that are consistent with $\A$:s observed values of these features. This implies that $\P$ belongs to a family of Gibbs distributions. 

Novelty (ii) also has relevance for (iii) since feature extraction is commonly used for data reduction within statistical learning; see, e.g., \cite[Section 5.3]{HastieTibshiraniFriedman2009}. But, as a consequence of the data processing inequality, this potentially implies a loss of information, regardless of how large the data set used to form $\A$'s beliefs about the values of the features is \cite[Section 2.8]{CoverThomas2006}, \cite[Problem 2.1]{DevroyeGyorfiLugosi1996}. Therefore, the Gibbs distribution beliefs of $\A$ about the value of $x_0$ are limited by which features are selected in the first place. We give a number of examples of how this provides fundamental limits in terms of LKA. The concept of secondary learning (iv) refers to the learning process of another agent $\tilde{\A}$ who lacks primary data $\DD$ but, on the other hand, uses other secondary data $\tilde{\DD}$ to learn how much $\A$ learned and acquired knowledge about $p$. In other words, $\tilde\A$ learns and acquires knowledge about $\P$ ($\A$ learns about $\A$'s learning) but not necessarily about $p$. This also has an impact on (iii) since machine learning algorithms often recapitulate the beliefs of humans, thereby performing secondary (rather than primary) LKA. We also demonstrate that the long-term effects of secondary learning are very similar to those of synthetic primary learning, whereby a third agent $\A^\prime$ learns from synthetic primary data $\DD^\prime$ generated by $\A$. 

\subsection{Organization of article}

Our paper is organized as follows. Section \ref{Sec:LKA} defines what it means that agent $\A$ has learned whether a proposition is true or not and whether he acquired knowledge about the proposition or not. Section \ref{Sec:Posterior} introduces a general framework for choosing the posterior distribution $\P$ as a Gibbs distribution that maximizes the entropy relative to $\P_0$, given side constraints that data $\DD$ provide. The concepts of Sections \ref{Sec:LKA} and \ref{Sec:Posterior}  are applied to LKA for feature-like data and Gibbs distributions in Section \ref{Sec:LKAGibbs} and to secondary learning in Section \ref{Sec:LKAGibbsSec}. Section \ref{Sec:Disc} provides a discussion and several proposed extensions. Finally, mathematical proofs and some additional examples are presented in the Supplement to this article \cite{DiazEtAlSupp2025}.

\section{Learning and knowledge}\lb{Sec:LKA}

In this section, we reproduce the definitions of LKA in \cite{HossjerDiazRao2022}. We also elaborate on the concept of discernment, proving some new results (Proposition 2.1 and Theorem \ref{FLFK}). Suppose that we have a measurable space $(\X,\F)$, where $\X$ is the set of possible worlds defined by the space of parameters $\X$ (i.e., each parameter value $x \in \X$ defines a world), whereas $\F$ is a $\s$-field on this set. It is assumed that $x_0\in\X$ represents the true world, whereas $\{x_0\}^c=\X\setminus \{x_0\}$ is a collection of counterfactuals. For a given proposition $p$, we define a measurable truth function $f_p: \X \to \{0,1\}$ s.t.
\begin{align}\label{TruthFunction}
	f_p(x) =
	\begin{cases}
		1 & \text{ if $p$ is true in the world $x$}, \\
		0 & \text{ if $p$ is false in the world $x$}.
	\end{cases}
\end{align}
Our goal is to learn $f_p(x_0)$, the truth value of the proposition in the true world. To accomplish this, we define the set 
\beq
	\TT = \{x \in \X : f_p(x) =1 \} \in \F
\lb{TT}
\eeq
of worlds in which $p$ is true. %As mentioned in Section \ref{Sec:Intro}, the mathematical formalization is ontologically frequentist and epistemologically Bayesian. 
The assumption that $p$ is either true or false in the true world ($f_p(x_0) \in \{0,1\}$) is aligned with a frequentist understanding of $f_p(x_0)$. 

\subsection{Discernment and belief}

We consider a Polish metric space $(\X, \F,d)$, i.e., a topological space $(\X,\mathcal O)$ such that $\F = \sigma(\mathcal O)$ is the Borel $\s$-field for the collection $\mathcal O$ of open sets of $\X$, and that $\X$ is complete with respect to the metric $d$. An agent $\A$ will assign its belief about $x_0$ according to a probability measure $\P$, whereas an ignorant agent $\I$ will assign its belief about $x_0$ following a probability measure $\P_0$. Thus, $\P$ and $\P_0$ are the respective predictors of $\A$ and $\I$ for $x_0$, the value of the true world. We refer to $P_0(x)$ and $P(x)$ as densities of $\P_0$ and $\P$ respectively, regardless of whether the corresponding probability measures are absolutely continuous, discrete or a mixture of both. Agents $\I$ and $\A$ will assign proba\-bilities to each $\AA \in \F$ by integrating over $\AA$ their density functions $P_0(x)$ and $P(x)$. That is, $\A$'s beliefs about $\AA$ are based on some data 
$
\DD \in \Delta
%\lb{D}
$
that $\I$ does not possess, where $\Delta$ is the set of all possible datasets, and computed from the posterior distribution as
\begin{align}\label{PA}
	\P(\AA) = \int_\AA P(x) \D x = 
	\frac{L(\DD \mid \AA)\P_0(\AA)}{L(\DD)},
\end{align}
%where $\D x$ is the Lebesgue measure $\nu(\D x)$ of a Euclidean space if $\X$ is an open subset of this space and $\P$ is absolutely continuous, whereas $\D x$ is the counting measure if $\X$ is finite or countable 
where $\P$ is absolutely continuous with respect to (wrt) the Lebesgue measure $\D x = \nu(\D x)$ if $\X$ is Euclidean or wrt the counting measure if $\X$ is countable. Moreover, $L(\DD \mid \AA) = \int_\AA L(\DD|x)P_0(x)\D x/\P_0(\AA)$ 
is the average likelihood of the parameters $x\in \AA$ for data $\DD$, whereas $L(\DD)=\int_\X L(\DD|x)P_0(x)\D x$ quantifies the overall strength of evidence $\DD$, from the perspective of $\I$ %$= \int_\DD \LL(\delta) \D \delta$ is the evidence. %, for $\LL(\delta)$ as defined in \eqref{Marginals}. 
%That is, there is a distribution $\QQ^*$ on the product space $\X\times\mathcal D$ such that $\P(X) \defeq \QQ^*(X \mid \DD)$, so that the averages in \eqref{muiP} become
%$$
%	\mu_i(\P) = \E_{\P}f_i(X) = \E_{\QQ^*(\cdot \mid \DD)}f_i(X) = \mu_i\left[\QQ^\ast(\cdot \mid \DD)\right].
%$$ 
(the Supplement \cite{DiazEtAlSupp2025}, provides the complete derivation of the posterior).
%In more detail, we assume that there is a random variable $D$ taking values on some measurable space $(\Delta, \mathcal D)$. For some underlying sample space $\Omega$, we define the random element $(X, D): \Omega \to \X \times \Delta$ that is $(\F \times \mathcal D)$-measurable. Moreover, to the measurable product space $(\X \times \Delta, \F \times \mathcal D)$ we associate a joint law $\QQ^*$ with density $Q^*(x, \delta) = P_0(x) L(\delta \mid x)$ and marginal densities
%\begin{align}\lb{Marginals}
%	\int_{\X} Q^*(x, \delta)\D x = L(\delta), \quad\quad
%	\int_{\Delta} Q^*(x, \delta)\D\delta = P_0(x).
%\end{align}
%Thus, the beliefs of $\I$ correspond to the density of $X$, whereas the posterior beliefs of agent $\A$ are obtained as the conditional density of $X$ given the event $\{D =\DD\}$, %With some slight abuse of notation, we call this density
%expressed as
%\begin{align}\lb{Post}
%	P(x)  \defeq Q^*(x \mid \DD) 
%	= \frac{Q^*(x, \DD)}{\int_\X Q^*(y, \DD)\D y}.
%\end{align}
%Finally, the belief \eqref{PA} of agent $\A$ about $\AA$ is obtained by integrating \eqref{Post} over $x\in \AA$, 
%\begin{align}\label{PA2}
%	\P(\AA) = \int_\AA \P(x) \D x = \frac{\mathbf L(\DD \mid \AA)\P_0(\AA)}{\mathbf L(\DD)},
%\end{align}
%Section \ref{Sec:Posterior} will focus on the properties of $\P$. 
%where $\LL(\DD \mid \AA) = \int_\AA \LL(\DD|x)\P_0(x)\D x/\P_0(\AA)$  is the average likelihood of the parameters $x\in \AA$ given the data $\DD$, and $\mathbf L(\DD) = \int_\DD \LL(\delta) \D \delta$ is the evidence, for $\LL(\delta)$ as defined in \eqref{Marginals}.
The densities $P_0(x)$ and $P(x)$ are measurable wrt $\s$-fields $\G_\I$ and $\G_\A$, respectively, with $\G_\I \subset \G_\A \subset \F$. This means that the beliefs of $\A$ and $\I$ are restricted to the information in $\G_\A$ and $\G_\I$, respectively. If $\G_\A = \s(\AA_1,\AA_2, \ldots)$
%\beq
%	\G_\A = \s(\AA_1,\AA_2, \ldots)
%	\lb{GPartition}
%\eeq
is generated by a countable partition $\PP=\{\AA_1, \AA_2, \ldots\}$
%\beq
%	\PP=\{\AA_1, \AA_2, \ldots\}
%	\lb{GPartition2}
%\eeq
of $\X$, it follows that the density 
\beq
P(x) = \sum_i p_i \1_{\AA_i}(x)
\lb{PPiecewise2}
\eeq
of $\P$ is piecewise constant over, and hence measurable wrt, the sets in $\PP$ that generate $\G_A$. Similarly, it follows that the density $P_0$ of $\P_0$ is piecewise constant over the sets of a partition $\PP_0$ (with $\sigma(\PP_0)=\G_\I$) that is coarser than $\PP$. The assumption that $\A$ is able to discern from a finer partition $\PP$ of $\X$ is natural, as it is often the case that refined experiments induce finer $\s$-fields for the potential resolution that data $\DD$ can provide about $x\in\X$. This is particularly obvious in the most extreme case, when $\I$'s discernment is the trivial $\s$-field $\G_\I=\{\X, \emptyset\}$. %, generated by a partition $\PP_0=\{\X\}$, and given by
%\beq
%	\G_\I=\{\X, \emptyset\}.
%	\lb{GI}
%\eeq
In particular, if $\G_\I=\{\X, \emptyset\}$ and $\X$ is bounded, then $\P_0$ has a constant density function over $\X$, making it necessarily the uniform distribution $\P_0(\AA) = |\AA|/|\X|$ 
%\beq
%	\P_0(\AA) = %\int_\AA \P_0(x)\D x = 
%	\frac{|\AA|}{|\X|}
%\lb{P0Unif}
%\eeq
for all $\AA\in\F$, where $|\X|$ refers to the number of elements of $\X$ for a finite set, or the Lebesgue measure $|\X|=\nu(\X)$ when $\X$ is a bounded subset of Euclidean space. Such a belief of $\I$ corresponds to a maximum entropy (maxent) distribution $\P_0$ over $\X$, and it represents a maximum state of ignorance.
%Note in particular that \eqref{P0Uniform} is identical to the uniform distribution \eqref{P0Unif}.

By construction of $\G_\A$, $\A$ has no advantage over $\I$ in terms of discerning how the probability is distributed {\it inside} the sets $\AA_i$ that generate $\G_\A$. On the other hand, if $\G_\A = \F$, there is maximum flexibility in the choice of $\P$. Therefore, the $\s$-fields generated by countable partitions of $\X$ represent upper limits for how much $\A$ and $\I$ are able to discern between the different worlds in $\X$. We formalize this as follows.

\begin{definition}[Discernment]\lb{DefDis}
Let $\G_\A$ be generated by a countable partition of $\X$. We say that an agent $\A$ cannot discern an event beyond $\G_{\A}$ if the following holds: For any $\s$-field $\G$ that is generated by a countable partition of $\X$, with $\G_\I \subset \G_\A \subset \G \subset \F$, and any $\F$-measurable function $g$, 
	\begin{align}\label{Discernment}
		\E_\P\infdivx{g}{\G} = \E_{\P_0}\infdivx{g}{\G} \quad \text{a.s.}
	\end{align}
\qed
\end{definition}

That is, the statement that $\A$ is unable to discern elements of $\X$ beyond $\G_\A$, means that the conditional expectation function $x \mapsto \E_\P\infdivx{g(X)}{\G}(x)$ of agent $\A$ is the same as that of the ignorant agent $\I$. In particular, if $g(x) = \1_\AA(x) = \1\{x \in \AA\}$, then $\P\infdivx{\AA}{\G} = \P_0\infdivx{\AA}{\G}$. Proposition 2.1 of the Supplement \cite{DiazEtAlSupp2025} provides additional interpretations and consequences of Definition \ref{DefDis}. Moreover, Example 1 of the Supplement \cite{DiazEtAlSupp2025} shows that discernment according to Definition \ref{DefDis} cannot always be extended to a $\s$-field $\G_\A$ that is not generated from a countable partition.

\subsection{Learning and knowledge acquisition}

We now formulate LKA in terms of active information (AIN). %Learning of proposition $p$ is defined as follows.

\begin{definition}\label{D:L} 
There is {\bf learning} of agent $\A$ about $p$, compared to an ignorant person $\I$, if the following condition holds:
\begin{enumerate}[label=\textbf{K\arabic*}, nosep]
 \item\label{Ka1}
 The active information (\ref{AIN}) of $\A$ relative to $\I$, for the set $\TT$ of worlds (\ref{TT}) for which $p$ is true, satisfies            
            \begin{align}\label{Learning}
		\left\{
			\begin{array}{ll}
				0 < I^+(\TT) \mbox{ and $p$ is true in the true world $x_0$},\\
				0> I^+(\TT)  \mbox{ and $p$ is false in the true world $x_0$}.
			\end{array}
		\right.
	\end{align}	
\end{enumerate}
	There is {\bf full learning} for $\A$ about $p$ (regardless of the beliefs of the ignorant person) if either $x_0 \in \TT$ and $\P(\TT)=1$, or if $x_0 \notin \TT$ and $\P(\TT)=0$.  
\end{definition}

\begin{remark}\label{R:L} 
	In words, $\A$ has learned about proposition $p$, compared to an ignorant agent $\I$, either when $p$ is true and $\A$'s belief about $p$ is higher than $\I$'s, or when $p$ is false and $\A$'s belief about $p$ is smaller than $\I$'s. Hence, it is possible for $\A$ to learn about true or false propositions. Thus, learning in the sense of \ref{Ka1} generalizes learning in the sense of \ref{K1}-\ref{K2}, since the latter only applies to true propositions. Agent $\A$ has fully learned $p$ if his beliefs about $p$ is 1 when $p$ is true or 0 when $p$ is false. 
\qed
\end{remark}

The notion of learning a proposition is limi\-ted, as it does not necessarily entail a particular belief about the true world. Therefore, it does not satisfy the conditions of a {\it justified} true belief, which requires having a belief {\it for the right reasons}. Knowledge acquisition is defined to cover this gap as follows: Whereas learning \ref{Ka1} is determining whether the given proposition $p$ is true or false, acquisition of knowledge about $p$ additionally requires a more confident estimate of the true world, in order to avoid learning $p$ with a wrong world model (by luck for instance). For this reason we need to augment \ref{Ka1} with two other conditions \ref{Ka2}-\ref{Ka3} in our definition of knowledge acquisition.

\begin{definition}\label{D:K}
	Agent $\A$ has acquired {\bf knowledge} about $p$, compared to an ignorant person $\I$, if $\A$ has learned about $p$ (condition \ref{Ka1} of Definition \ref{D:L} holds), and additionally the following two conditions hold: 
	\begin{enumerate}[label=\textbf{K\arabic*}, nosep]
            \setcounter{enumi}{1}
%    		\item\label{Ka1} The criteria of \eqref{Learning} in Definition \ref{D:L} are satisfied. 
                \item\label{Ka2} $x_0 \in \text{supp}(\P)$, the support of $\P$.
    		\item\label{Ka3} For all $\epsilon > 0$, the closed ball $B_\epsilon [x_0] \defeq \{x \in \X: d(x, x_0) \le \epsilon\}$ is such 			that $I^+(B_\epsilon[x_0])\ge0$, with strict inequality for some $\epsilon>0$, where $d$ is a metric over $\X$.
	\end{enumerate}
	Agent $\A$ has acquired {\bf full knowledge} about $p$ (regardless of the beliefs of the ignorant person) if 				$\P=\bde_{x_0}$, the point mass at $x_0$.
\qed
\end{definition}

Condition \ref{Ka1} ensures that $\A$ must learn about $p$ to acquire knowledge; therefore, KA is a more stringent concept than learning, as illustrated by Figure \ref{LvK}. Condition \ref{Ka2} says that the true world $x_0$ is among the pool of possibilities for $\A$, which is formally equivalent to saying that $\A$ has a positive belief for every open ball centered at $x_0$ (that is, if for all $\epsilon > 0$, $\P(B_\epsilon(x_0))>0$, where 
\beq
B_\epsilon(x_0) \defeq \{x \in \X: d(x, x_0) < \epsilon\}
\lb{Bepsx0}
\eeq
is the open ball of radius $\epsilon$ centered at $x_0$). This in turn explains \ref{Ka3}, that the belief in $x_0$ under $\P$ is stronger than that under $\P_0$, i.e., that the beliefs of $\A$ are more concentrated around $x_0$ than those of $\I$. 

\begin{remark}
Note that \ref{Ka2} is implied by \ref{Ka3} (and hence is obsolete) when $x_0\in\mbox{supp}(\P_0)$. This includes, for instance, the case when $\X$ is bounded or finite, and $\P_0$ is the uniform distribution on $\X$. On the other hand, \ref{Ka3} is satisfied but not \ref{Ka2} when $\X=[0,1]$, $x_0=0.75$, $P_0(x)=2 \cdot \1_{[0,0.5]}(x)$ and $P(x) = 4x \cdot \1_{[0,0.5]}(x)$. Conditions \ref{Ka2}-\ref{Ka3} can also be used as a definition for acquiring knowledge about $x_0$. This is weaker than acquiring knowledge about $p$, since the latter requires increased/decreased beliefs in $p$ when $p$ is true/false, and justification in terms of increased knowledge about $x_0$. Consider for instance the following example suggested by a reviewer: $\X=[0,1]$, $\TT=[0.4,0.6]$, $x_0=0.6$, $P_0(x)=1$, and $P(x)=5\cdot \1_{[0.59,0.79]}(x)$. In this case, $p$ is true and $x_0$ is at the boundary of $\TT$. It can be seen that \ref{Ka3} is satisfied but not \ref{Ka1}. This is to say that agent $\A$ has sacrificed knowledge about $p$ in order to attain knowledge about $x_0$. However, it is possible for $\A$ to attain knowledge about $p$, for instance by having $P(x)=5 \cdot \1_\TT(x)$. 
\qed
\end{remark}

\begin{figure}[t]
	\includegraphics[scale=0.6]{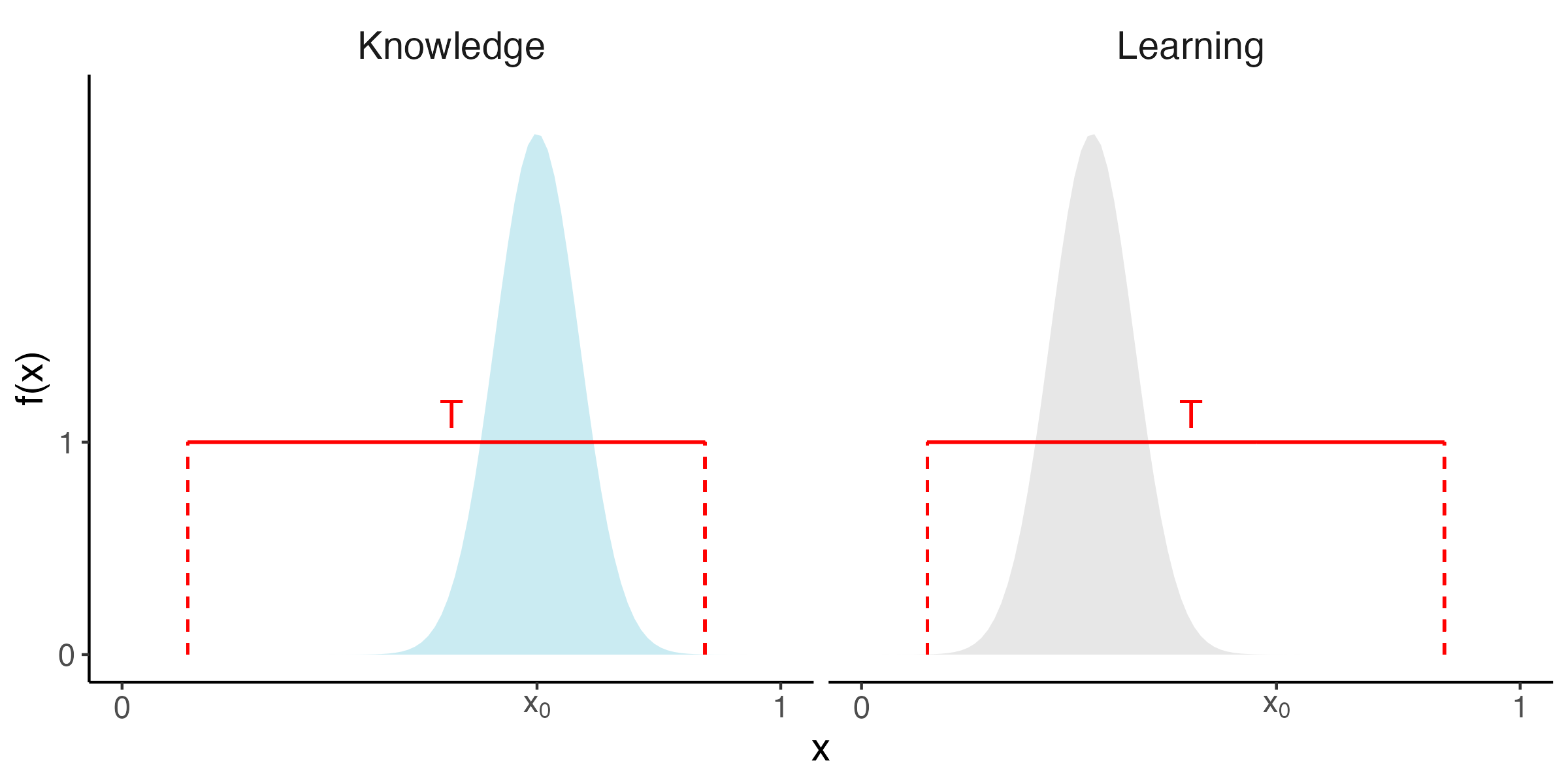}
	\caption{Learning versus KA: The set of possible worlds is $\X=[0,1]$, the set of worlds where a given proposition $p$ 	is true is given by $\TT$, the true world is $x_0$, and $\P_0$ is the uniform measure on $\X$. Thus $\P_0(\TT) = \textrm{length}(\TT) 	< 1$. The light blue region in the LHS represents the beliefs of an agent $\A_1$, whereas the gray region in the RHS represents the beliefs of another agent $\A_2$. Since the beliefs of the two agents are fully concentrated in $\TT$, $\P_{\A_1}	(\TT) = \P_{\A_2}(\TT) = 1$. Therefore, the two agents fully learned about $p$. However, since in the RHS $x_0           	\notin \textrm{supp}(\P_{\A_2})$, $\A_2$ does not acquire knowledge, whereas $\A_1$ does as his beliefs are 	more concentrated around $x_0$ than those of the ignorant agent with belief $\P_0$. Nonetheless, full KA is not possible for $\A_1$ as $\P_{\A_1}$ is continuous.}\lb{LvK}
\end{figure}

Our next result details how the discernment $\G_\A$ of agent $\A$ sets limits to his ability to learn various propositions $p$, with different truth sets $\TT$. In more detail, we provide sufficient conditions on $\P_0$, $\G_\A$ and $\TT$ for not having full learning (\ref{FuLeT} and \ref{FuLeF}) and not having full KA (\ref{FuKn}) respectively. Moreover, we provide sufficient conditions on $\P_0$, $\G_\A$ and $\TT$ for obtaining full learning (\ref{FuLeTT} and \ref{FuLeFF}) and full KA (\ref{FuKnn}) respectively. In all cases, this is {\sl regardless} of the type of data $\DD$ that $\A$ receives within his resolution $\G_\A$. In particular, conditions \ref{FuLeT} and \ref{FuLeF} for not having full learning are such that the truth function $f_p=\1_\TT$ of $p$ is not $\G_\A$-measurable. 

\begin{theorem}\lb{FLFK}
    Consider the Polish space $(\X, \F, d)$, where $\F = \s(\mathcal O)$. Let $\P_0$ be a %fully supported 
    probability measure on $(\X, \F)$ and define another probability measure $\P$ on $(\X, \F)$ as in \eqref{PA}, where $\P_0$ and $\P$ represent beliefs about the true world $x_0\in\X$ of two agents $\I$ and $\A$ respectively. Assume that their respective densities $P_0(x)$ and $P(x)$ are measurable wrt $\s$-fields $\G_\I$ and $\G_\A$ on 	$\X$, with $\G_\I \subsetneq \G_\A \subset \F$. Assume further that $\G_\A = \s(\PP)$ is generated from a countable partition $\PP$, such that $\P_0(\AA_i) > 0$ for all $\AA_i \in \PP$ and none of the $\AA_i \in \PP$ is $\G_\I$-measurable Let $p$ be a proposition that is true in a set of worlds $\TT \in \F$, defined in \eqref{TT}. Then
	\begin{enumerate}[nosep, label = \roman*.]
		\item\lb{FuLeT} If for all $\AA \in \PP$, it holds that $\AA \not \subset \TT$ and $\P_0(\AA \setminus \TT) >0$, then $\P(\TT) < 1$. In particular, if $p$ is true in the true world $x_0$, full learning of $p$ is not possible.  
		\item\lb{FuLeTT} Suppose \ref{FuLeT} fails in the sense that there is an $\AA \in \PP$ such that $\AA \subset \TT$. Then we can choose $x_0$ so that $p$ is true in $x_0$, and $\P$ according to \eqref{PPiecewise2}, so that %there is full learning of $p$, i.e.\ 
        $\P(\TT) = 1$.
		\item\lb{FuLeF} If for all $\AA \in \PP$, it holds that $\TT\cap\AA\ne\emptyset$ and $\P_0(\TT\cap\AA) >0$, then $\P(\TT) >0$. In particular, if $p$ is false in the true world $x_0$, full learning of $p$ is not possible. 
		\item\lb{FuLeFF} Suppose \ref{FuLeF} fails in the sense that there is $\AA \in \PP$ such that $\AA \cap \TT = \emptyset$. Then we can choose $x_0$ such that $p$ is false in $x_0$, and $\P$ according to \eqref{PPiecewise2}, so that %there is full learning of $p$, i.e.\ 
        $\P(\TT) = 0$.
		\item\lb{FuKn} If there is $\AA \in \PP$ such that $\{x_0\} \subsetneq \AA$ and $\P_0(\AA \setminus \{x_0\}) > 0$, then $\P(\{x_0\}) < 1$ and full KA is not possible.
		\item\lb{FuKnn} If $\{x_0\} \in \PP$, then it is possible to choose $\P$ according to \eqref{PPiecewise2} such that $\P(x_0) = 1$.
	\end{enumerate}
\end{theorem}

%\textcolor{red}{It seems to me that Theorem \ref{FLFK} can be stated without assuming that $\P$ and $\P_0$ have densities. Anyhow, the first point of the remark below explains what happens if the Radon-Nykodim derivatives exist. OH: Equation (7) can be interpreted more generally, allowing some sets of the partition to be single points, as in the last part vi of Theorem 2.4. }

\begin{remark}
	The conditions imposed in Theorem \ref{FLFK} are, in general, easy to obtain, and the result is true with great generality. Note in particular the following: 
	\begin{itemize}[nosep]
	 	\item 
        $\G_\I=\s(\PP_0)$ is generated from a partition $\PP_0$ coarser than $\PP$, with $\P_0(\AA)>0$ for all $\AA\in\PP_0$. Since $\P_0$ is measurable wrt $\G_\I$, the conditional distribution of $\P_0$ is uniform over all $\AA\in\PP_0$. This implies that the conditional distribution of $\P_0$ is uniform over all sets $\AA\in\PP$ of the finer partition as well. 
     %   \item
	%	The requirement that $\P_0$ is fully supported on $(\X, \F)$ is natural. A way to see this is by making explicit that $\P_0$ is absolutely continuous with respect to another measure $\nu$, i.e., $\P_0 \ll \nu$. If $\nu$ is the Lebesgue measure, then for every open set $\mathsf O \in \mathcal O$, $\nu(\mathsf       O) > 0$. Since $\P$ as defined in \eqref{PA} is also absolutely continuous with respect to $\P_0$ (so $\P \ll \P_0$), $\P(\mathsf O) = 0$ if $\P_0(\mathsf       O) = 0$, which means that no agent can learn anything he has ruled out from the beginning. If $\P_0$ is not fully supported on $(\X, \F)$, then we can find      a set $\mathsf O_1$ $\in \mathcal O$ with $\nu(\mathsf O_1)>0$ such that $0=\P_0(\mathsf O_1) = \P(\mathsf O_1)$. This would imply that $x_0\in\mathsf O_1$      was ruled out from the start for agent $\A$, for a set $\mathsf O_1$ with positive Lebesgue measure, regardless of what the data $\DD$ tell him. This is a restriction unless the beliefs of agent $\I$ are based on {\sl previous} data that have excluded the possibility $x_0\in\mathsf O_1$. 
	%	\item 
         %   A corollary of the previous item is that if $\P_0$ is absolutely continuous with respect to some measure $\nu$ and $\nu(\{x_0\}) = 0$, then full knowledge is not attainable.
            \item 		
		Suppose $\X = \R$, $\AA =  [a,b]\in\PP$, $\TT = (a,b)$, and make $\P_0$ absolutely continuous wrt the Lebesgue measure on $\R$. Then, full learning can be obtained in Theorem \ref{FLFK}.\ref{FuLeT} even if $\TT \subset 		\AA$. Thus the requirement that $\P_0(\AA \setminus \TT) >0$ for all $\AA\in\P$. \qed
	\end{itemize}
\end{remark}

\section{Maximum entropy and Gibbs posterior distributions}\lb{Sec:Posterior}

\subsection{Default choice of posterior}

We will construct the posterior distribution $\P$ in \eqref{PA} from the prior distribution $\P_0$, using a set $\mathbf f = (f_1,\ldots, f_n)$ of $n$ feature functions $f_i:\X\to\R$, $i=1,\ldots,n$, with $f_i(X)$ the value of feature $i$ for some randomly generated $X\in\X$. Moreover, $\P$ is generated from $\P_0$ in such a way that outcomes in regions of $\X$ where $f_i$ is large are either more or less likely under $\P$ compared to $\P_0$, given that the other $n-1$ features do not change. In more detail, define $\QQQ$ as the set of probability measures on $\X$. For any $\QQ\in\QQQ$, let   
\beq
	\mu_i(\QQ) = \E_{\QQ}f_i(X)
	\lb{muiP}
\eeq
represent the expected value of feature $i=1,\ldots,n$ under $\QQ$, and denote the corresponding vector of expected features as $\bmu(\QQ)= (\mu_1(\QQ),\ldots,\mu_n(\QQ))$. For any vector $\bmu=(\mu_1,\ldots,\mu_n)$ of expected features, let
\beq
\P = \P_{\bmu} = \mbox{arg} \inf_{\QQ\in\QQQ(\bmu)} \infdiv{\QQ}{\P_0} 	
\lb{Popt}
\eeq
be the distribution that minimizes the Kullback-Leibler divergence $\infdiv{\QQ}{\P_0} = \E_{\QQ}\log[Q(X)/P_0(X)]$ (or equivalently maximizes the entropy relative to $\P_0$) among all probability distributions $\QQ\in\QQQ(\bmu)$, that is, all probability measures that firstly satisfy $\QQ\in\QQQ$, and secondly  
\beq
	\bmu(\QQ) = \bmu. 
	\lb{muiQP}
\eeq
In the Supplement \cite{DiazEtAlSupp2025}, we motivate that the solution to the constrained minimization problem \eqref{Popt}--\eqref{muiQP} is the Gibbs distribution $\P=\P_{\bmu}$ with density
\begin{align}\label{Gibbs2}
P(x) = P_{\bmu}(x) = Q_{\bla}(x) = \frac{P_0(x) e^{\bla \cdot \mathbf f(x)}}{Z_{\bla}},
\end{align}
where $\bla = (\lambda_1, \ldots, \lambda_n) = \boldsymbol\lambda(\bmu) \in \R^n$ is a vector of dimension $n$ chosen so that \eqref{muiQP} holds, and $Z_{\boldsymbol\lambda}$ is a normalizing constant that makes $\QQ_{\boldsymbol{\lambda}}$ a probability measure.
%\beq
%	Z_{\boldsymbol\lambda} = \int_{\X} \P_0(x) e^{\boldsymbol{\lambda} \cdot \mathbf f(x)}\D x = \E_{\P_0}e^{\boldsymbol{\lambda} \cdot \mathbf f(X)}
%	\lb{Zlambda}
%\eeq
%is a normalizing constant, selected so that $\QQ_{\boldsymbol{\lambda}}$ is a probability measure. In \eqref{Zlambda}, we interpret $\D x=\nu(\D x)$ as the Lebesgue measure when $\X$ is a subset of a Euclidean space or as the counting measure when $\X$ is a finite set \textcolor{red}{(We said this before, not for Gibbs)}. 
Let $\DD \in \Delta$ be a data set available to agent $\A$ that is informative for the values of the $n$ features. We will apply (\ref{Gibbs2}), with $\hmu_i=\hmu_i(\DD)$ the features observed or estimated by $\A$, that are functions of data $\DD$, and $\hbmu(\DD)$ the corresponding vector of observed features. With this choice of $\bmu$, we may interpret the Gibbs distribution $\QQ_{\boldsymbol{\lambda}}$ in (\ref{Gibbs2}) as a posterior distribution of agent $\A$ with density
\begin{align}\label{Bayes2}
	\begin{aligned}
		P(x) = P_{\hbmu(\DD)}(x) = L(\DD \mid x) P_0(x)/L(\DD)
	\end{aligned}
\end{align}
when the prior distribution is $\P_0$ and the likelihood is %\giventhat{f(x)=\frac{x^2}{2}}{x=1,2,3,\dotsc}
\beq
	L\giventhat{\DD}{x} = e^{\boldsymbol{\lambda} \cdot \mathbf f(x)}.
\lb{LikGibbs}
\eeq
The connection between Gibbs distributions and Bayesian statistics has been exploited in high-dimensional statistics and statistical physics \cite{Barbier2020, ZdeborovaKrzakala2016}. %When $\X$ is finite or bounded, it is natural to impose a maxent prior $\P_0$ on $\X$, equal to the uniform distribution \eqref{P0Unif}. 
Note that the formal likelihood in \eqref{LikGibbs} is proportional to a member of an exponential family with parameter $x\in\X$ and sufficient statistic $\bla=\bla(\hbmu(\DD))$ \cite{LehmannCasella1998}. In particular, $x$ is a natural parameter of this family if $x=\mathbf f(x)$. However, \eqref{LikGibbs} is not necessarily an actual likelihood, since 
$$
\int_\Delta L(\delta|x) \D\delta = 
\int_\Delta  e^{\bla(\hmu(\delta)) \cdot \mathbf f(x)}\D\delta
$$
is typically different from 1. The vector $\bla=\bla(\hbmu(\DD))$ of the formal likelihood in \eqref{LikGibbs} will be chosen to be consistent with the constraints \eqref{muiQP} of the optimization problem \eqref{Popt} that data $\DD$ provide. Since \eqref{LikGibbs} is not the true likelihood of data $\DD$, we refer to $P(x)$ in \eqref{Gibbs2} as a quasi-posterior distribution, obtained by inserting \eqref{LikGibbs} into \eqref{Bayes2}. 
%Recalling from \eqref{Post} that there is a distribution $\QQ^*$ on the product space $\X\times\mathcal D$ such that $\P(X) \defeq \QQ^*(X \mid \DD)$, it follows from \eqref{muiP} that these constraints are
%$$
%	\mu_i(\P) = \E_{\P}f_i(X) = \E_{\QQ^*(\cdot \mid \DD)}f_i(X) = \mu_i\left[\QQ^\ast(\cdot \mid \DD)\right].
%$$ 
Suppose data $\DD=(\DD_1,\ldots,\DD_N)$
%\beq
%\DD=(\DD_1,\ldots,\DD_N)
%\lb{DN}
%\eeq
of size $N$ is an observation of the random vector $D=(D_1,\ldots, D_N)$. For instance, the components $D_k$ of $D$ could be iid variables. The following proposition concerns the asymptotic posterior distribution $\P=\P_{\hbmu(D)}$ as $N$ gets large: 

\begin{proposition}\lb{Prop:PAsympt}
Let $\P=\P_{\hbmu(\DD)}$ refer to the solution of the optimization problem \eqref{Popt}, with an estimated feature vector $\hbmu=\hbmu(\DD)$ that is an observation of the random vector $\hbmu(D)$. Assume that convergence in probability
\beq
\hbmu(D)\stackrel{p}{\to} \f(x_0)
\lb{muConvLargeN}
\eeq
holds as $N\to\infty$, for data $D=(D_1,\ldots,D_N)$, where $x_0$ is the true but unknown value of $x$. Then  $\P = \P_{\hbmu(D)} \stackrel{\mathcal L}{\to} \P_\infty$ converges weakly to $\P_\infty$,
%\beq
%\P(D) \stackrel{\mathcal L}{\to} \P_\infty
%\lb{PConvLargeN}
%\eeq
as $N\to\infty$ a.s., where $\P_\infty$ is the Gibbs distribution \eqref{Gibbs2} with $\bmu(\P_\infty)=\f (x_0)$.
\end{proposition}

Although Proposition \ref{Prop:PAsympt} is mathematically simple, it is a key result to understand the limits of asymptotic knowledge acquisition. The proposition states that $\P_\infty$ is the asymptotic limit of the posterior $\P$ as $N\to\infty$. In Section \ref{Sec:LKAGibbs}, we will find that $\P_\infty$ differs from a point mass $\bde_{x_0}$ at the true world $x_0$ when the number of features $n$ is too small. In view of Definition \ref{D:K}, this is to say that it is not possible to have full KA asymptotically as $N\to\infty$, unless the number of features is large enough.  The following theorem shows that it is possible to obtain a $1/\sqrt{N}$ rate of convergence of $\P_{\hbmu(D)}$ towards $\P_\infty$, when $\hbmu(D)$ is an unbiased sample average, regardless of whether $\P_\infty$ equals $\bde_{x_0}$ or not: 

\begin{theorem}\lb{Thrm:PAsympt}
Assume that the estimates features $\hbmu(\DD)$ are obtained from an independent sample $\DD=(\DD_1,\ldots,\DD_N)$ as a sample average
\beq
\hbmu = \hbmu(\DD) = \frac{1}{N} \sum_{k=1}^N \hbmu(\DD_k)
\lb{muDDInd}
\eeq
where $\{\hbmu(\DD_k)\}_{k=1}^N$ are observations of $\{\hbmu(D_k)\}_{k=1}^N$. Assume that the $\hbmu(D_k)$ are iid, unbiased ($E[\hbmu(D_k)]=\f(x_0)$) and that $\Var[(\hbmu(D_k)]=\bSi$, where $\bSi$ is a covariance matrix of order $n$. Then
\beq
\sqrt{N}(\hbmu(D)-\f(x_0)) \stackrel{\mathcal L}{\to} N(0,\bSi)
\lb{muDDConv}
\eeq
as $N\to\infty$. In addition
\beq
\sqrt{N}(\P_{\hbmu(D)}-\P_\infty) \stackrel{\mathcal L}{\to} \W
\lb{PDDConv}
\eeq
as $N\to\infty$ a.s., where $\P_{\hbmu(D)}$ and $\P_\infty$ are defined as in Proposition \ref{Prop:PAsympt}, whereas $\W$ is a Gaussian signed measure on $\X$, with $\W(\AA)\sim N(0,C(\AA,\AA))$ and $\Cov(\W(\AA),\W(\BB))=C(\AA,\BB)$ for all $\AA,\BB\in\F$, and with $C(\AA,\BB)$ defined in the proof.  
\end{theorem}

\begin{example}[Finite populations]\lb{Exa:Finite}
Suppose $\X=\{x_1,\ldots,x_d\}$ is a finite set. We can generate $\X$ from a population $\EE$ of (a large) size $M$, which is partitioned into $d$ nonempty subsets $\EE=\cup_{k=1}^d x_k$, corresponding to a partition $\X =\{x_1, \ldots, x_d \}$ of $\EE$. The measurable space $(\EE,\sigma(\X))$ consists of all $2^d$ finite unions of sets $x_i$, and a distribution $\QQ$ on $(\EE,\sigma(\X))$ corresponds to probabilities $q_k=Q(x_k)$ for $k=1,\ldots,d$. It belongs to the $(d-1)$-dimensional simplex $\QQQ \defeq \left\{(q_1, \ldots, q_d) \in (\R^+)^d : q_1 + \cdots + q_d = 1 \right\}$, where $\R^+$ is the set of nonnegative real numbers. Since $\X$ is finite, without further background information, we impose a uniform prior $\P_0 = \{p_{01}, \ldots, p_{0d}\}$ with $p_{0k} = 1/d$. The distribution
$\P = \{p_1, \ldots, p_d\} \in \QQQ$ that is in maxent relative to $\P_0$ is the Gibbs distribution with probability function 
$p_k = P_{\hbmu(D)}(x_k) = e^{\bla \cdot \mathbf f(x_k)}/\sum_{l=1}^d e^{\bla \cdot \mathbf f(x_l)}, \quad k=1,\ldots,d$, with $\bla=\bla(\hbmu(D))$ chosen so that the expected feature vector of $\P$ equals the observed features $\hbmu(D)$, cf.\ \eqref{muiQP}. Since the simp\-lex $\QQQ$ is $(d-1)$-dimensional, the number of features of the Gibbs distribution must satisfy $1\le n \le d-1$ in order to avoid overparametrization. We will return to Example \ref{Exa:Finite} in Section \ref{Sec:LimLKAFin}, in order to illustrate Proposition \ref{Prop:PAsympt} and how the number of features sets limits to asymptotic KA for data $D=(D_1,\ldots,D_N)$ as $N\to\infty$. 
\qed
\end{example}

\subsection{Biased choice of posterior}\lb{Sec:Biased}

The beliefs $\P$ of agent $\A$ are based on a posterior Gibbs distribution%\eqref{Gibbs2}
. It includes a prior $\P_0$ that is typically chosen to be maxent over $\X$%(like the uniform distribution \eqref{P0Unif} when $\X$ is bounded)
, and a likelihood $L(\DD|x) = e^{\bla \cdot\f(x)}$ %in (\ref{LikGibbs}) 
that makes the posterior $\P$ maxent relative to $\P_0$, given the observed features $\hbmu=\hbmu(\DD)$. Therefore, we regard the prior and the likelihood  of $\A$ as default. 

Consider another agent $\tilde{\A}$ who makes use of the same data $\DD$ as $\A$, but whose likelihood $\tilde{L}(\DD|x)$ and prior density $\tilde{P}_0(x)$ are possibly different from those of $\A$. We will regard the beliefs of $\tilde{\A}$, based on a posterior density
\beq
	\tilde{P}(x) = \tilde{L}(\DD \mid x) \tilde{P}_0(x)/\tilde{L}(\DD),
	\lb{tildeP}
\eeq
as biased in comparison to those of $\A$. Following \cite{Montanez2017b, MontanezEtAl2019, MontanezEtAl2021} to measure bias in algorithms, and \cite{DiazRao2021, HossjerEtAl2024, ZhouEtAl2023} to measure the bias of prevalence estimators of COVID-19, we use AIN to measure bias for the beliefs of $\tilde{\A}$, compared to those of $\A$. That is, for a target $\TT\in\X$ %we let 
\begin{align}
	\mbox{Bias}(\TT;\P,\tilde{\P}) = I^+(\TT;\P,\tilde{\P}) 
	= I^+(\TT;\P_0,\tilde{\P}) - I^+(\TT;\P_0,\P) 
	= \log [\tilde{\P}(\TT)/\P(\TT)]
	\lb{AINA}
\end{align}
refer to the change in AIN by considering $\tilde\P$ instead of $\P$. An instance of biased learning will be given in Example \ref{Ex:FiPo1} of Section \ref{Sec:LimLKAFin}. When only $\tilde{\A}$'s likelihood is misspecified as 
\beq
	\tilde{L}(\DD \mid x) = e^{\tilde{\bla} \cdot \mathbf f(x)}
	\lb{LikGibbs2}
\eeq
for some $\tilde{\bla}\ne\bla$, whereas the prior of $\tilde{\A}$ is the same as that of $\A$, it follows that 
\beq
	\mbox{Bias}(\TT;\bla,\tbla) = \log \frac{Z_{\tbla}(\TT)Z_{\bla}(\X)}{Z_{\bla}(\TT)Z_{\tbla}(\X)},
	\lb{AINA2}
\eeq
where $Z_{\bla}(\TT) = \int_{\TT} P_0(x) e^{\boldsymbol{\lambda} \cdot \mathbf f(x)} \D x$. In Section \ref{Sec:ML}, %we will use \eqref{AINA2} in another setting 
where $\tilde{\A}$ uses secondary data $\tilde{\DD}$ to learn about $\A$'s learning, \eqref{AINA2} will quantify the error in $\tilde{\A}$'s learning about $\A$'s learning. 

\section{LKA for Gibbs distributions}\lb{Sec:LKAGibbs}

We now combine Sections \ref{Sec:LKA} and \ref{Sec:Posterior} to consider LKA. A particular focus will be paid to whether full LKA is possible or not, for instance when $N\to\infty$. Although partial LKA is often good enough, it turns out that for many models, explicit conditions can be obtained for when full LKA is possible. As we will see, the quasi-Bayesian approach with Gibbs distributions is very powerful for finding limits of KA. The crucial question is whether the family of Gibbs posterior distributions in \eqref{Gibbs2} is rich enough when $\bla$ varies. Recall from Section \ref{Sec:Posterior} that $\A$ has a Gibbs posterior, based on $n$ feature functions $f_1,\ldots,f_n$ and data $\DD$ in terms of $\A$'s observed expected beliefs $\hbmu=\hbmu(\DD)$ about the values of the $n$ features. Since $\A$ forms his beliefs about $x_0$ based on the largeness/smallness of the feature functions, it is reasonable to define his discernment $\G_\A = \sigma(f_1,\ldots,f_n)$
%\beq
%	\G_\A = \sigma(f_1,\ldots,f_n)
%	\lb{GAGibbs}
%\eeq
as the smallest $\s$-field that makes all feature functions measurable. Indeed, for a uniform prior we deduce from \eqref{LikGibbs} that $\A$'s likelihood as well as his posterior density are both measurable wrt $\G_\A$. When the feature functions are binary indicator functions, this discernment is reduced to a finite partition $\G_\A = \s(\AA_1,\ldots,\AA_l)$, where $n\le l \le 2^n$ is the collection of non-empty intersections of the sets $\{f_i^{-1}(0),f_i^{-1}(1); i=1,\ldots,n\}$. In particular, $l=n$ when the sets $f_i^{-1}(1)$ form a finite partition of $\X$. 

From \eqref{Gibbs2}, each feature $i$ contributes to increase/decrease $\A$'s beliefs about $x_0$ in regions where $\lambda_i f_i(x)$ is large/small. This has an impact on learning about a proposition $p$ that is true whenever the value of feature $i$ is at least a constant value $f_0$. This corresponds to a truth function $f_p(x) = \1_\TT(x)$, with 
\beq
	\TT = \{x \in \X : f_i(x)\ge f_0 \}
	\lb{fpfi}
\eeq
the set of worlds in which $p$ is true. Proposition \ref{Prop:Learnfi} provides details about learning $p$.

\begin{proposition}\lb{Prop:Learnfi}
Consider a proposition $p$ which is true in the set of worlds \eqref{fpfi}, for some $i\in\{1,\ldots,n\}$. Assume further that
$\min_{x\in\X} f_i(x) \le f_0 \le \max_{x\in\X} f_i(x)$,
with at least one of the two inequalities being strict. Then $\P(\TT) = \QQ_{\bla}(\TT)$ is a strictly increasing function of $\lambda_i$, with 
\begin{align}\lb{PlambdaLimit}
    \lim_{\lambda_i\to -\infty} \QQ_{\bla}(\TT) = 0 \quad \textrm{and} \quad \lim_{\lambda_i\to \infty} \QQ_{\bla}(\TT) = 1 
\end{align}
%    \beq
%    \begin{array}{rcl}
%    \lim_{\lambda_i\to -\infty} \QQ_{\bla}(\TT) &=& 0,\\
%     \lim_{\lambda_i\to \infty} \QQ_{\bla}(\TT) &=& 1    \end{array}
%    \lb{PlambdaLimit}
%    \eeq
    when the other $n-1$ components of $\bla$ are kept fixed. 
    In particular, $\A$ learns $p$ (compared to $\I$), if the following two conditions hold:
	\begin{enumerate}[label=(\roman*)]
		\item $\lambda_j=0$ for all $j \in \{1,\ldots,n\} 	\setminus \{i\}$, 
		\item either $\lambda_i>0$ and $f_i(x_0)\ge f_0$, or $\lambda_i<0$ and $f_i(x_0)<f_0$.  
	\end{enumerate}     
\end{proposition}

In principle, by \eqref{PlambdaLimit}, it is possible for $\A$ to attain full learning about a proposition that is true when one feature exceeds a given threshold. It is enough in this case for $\A$ to have data $\DD$ that lead to the appropriate estimated features $\hbmu=\hbmu(\DD)$, and the corresponding sufficient statistic $\bla=\bla(\hbmu(\DD))$ of the likelihood \eqref{LikGibbs}, that make $\QQ_{\bla}(\TT)$ close to 1 (0) when $p$ is true (false). However, as it will be seen in Sections \ref{Sec:LimLKAFin}--\ref{Sec:LimDisc}, for other types of propositions, neither full learning nor full KA is guaranteed when the number of features is too small.  

\subsection{Fundamental limits of KA for classification on finite populations}\lb{Sec:LimLKAFin} % Added title and description below

This section presents examples of LKA  for classification over finite populations. Example \ref{Ex:FiPo1} illustrates with one binary feature that full knowledge might not be possible even if full learning is obtained. Theorem \ref{Prop:FinPop} generalizes the situation to multiple features, proving that there are fundamental limits for full KA.

\begin{example}[Finite populations with one binary feature.]\lb{Ex:FiPo1}
Continuing Example \ref{Exa:Finite}, recall that $\EE$ is a population with $M$ subjects, partitioned into $d$ subsets %(or subpopulations) 
(say, $d$ cities)
\beq
	\X=\{x_1,\ldots,x_d\}.
	\lb{Xfinite}
\eeq
%Assume that the objective of $\A$ is to find to which subset $x_0=x_{k_0}$ a particular subject $S$ belongs. 
Assume %further 
that the first $h$ cities $\mathsf N^c \defeq \{x_1,\ldots,x_h\}$ are southern, whereas the remaining $d-h$ cities  $\mathsf N \defeq \{ x_{h+1},\ldots,x_d\}$ are northern. Suppose the only feature function $f(x_k) = \1_{\mathsf N}(x_k)$ is an indicator as to whether a city is northern. Consider the proposition
$$
p: \mbox{Subject $\mathcal S$ resides in a northern city},
$$
and let $x_0=x_{k_0}$ be the city where $\mathcal S$ actually lives. The truth function \eqref{TruthFunction} of $p$ equals the feature function $f(x_k) = \1_{\mathsf N}(x_k)$, i.e., $f_p=f$, so the set of worlds for which $p$ is true is $\TT  =  \{x_{h+1},\ldots,x_d\} = \mathsf N$. 
Assume that,
%$\DD$ provides $\A$ with some information as to whether subject $\mathcal S$ resides in a northern subpopulation or not. More precisely
based on data $\DD$, $\A$ believes that, with probability $\hmu=\hmu(\DD)=\E_\P f(X)$, subject $\mathcal S$ lives in a northern city. The Gibbs distribution \eqref{Gibbs2}, with a uniform prior $P_0(x_k)=1/d$, simplifies to %a probability function 
\beq
	P(x_k) = 
	\left\{\begin{array}{ll}
		\frac{1}{h + (d-h)e^\lambda} = \frac{1-\hmu}{h}; & k=1,\ldots,h,\\
		\frac{e^\lambda}{h + (d-h)e^\lambda} = \frac{\hmu}{d-h}; & k=h+1,\ldots,d,
	\end{array}\right. 
\lb{PSubpop}
\eeq
whereas the $\s$-field $\G_\A = \{\emptyset, \mathsf N, \mathsf N^c, \X \}$. Suppose $p$ is true ($x_0\in \TT$). Then, KA requires more than learning if $d-h\ge 2$, since learning occurs when 
\beq
	\P(\TT) = P(x_{h+1})+\ldots + P(x_d) > (d-h)/d = \P_0(\TT),
	\lb{Ex1L}
\eeq
which, by Proposition \ref{Prop:Learnfi} with $n = i = f_0 =1$, is equivalent to $\lambda>0$. In particular, full learning is attained when the LHS of \eqref{Ex1L} equals $1$. However, defining the metric $d(x,y)=\1\{x\ne y\}$ on $\X$, Condition \ref{Ka3} of Definition \ref{D:K} implies that, on top of \eqref{Ex1L}, full KA is not possible when $d-h\ge 2$, because
\beq
	P(x_0) \le 1/(d-h) < 1.
	\lb{Ex1KA}
\eeq
Thus, KA requires more than learning when $x_0\in\TT$ and $d-h\ge 2$. In order to illustrate this asymptotically ($N\to\infty$), consider a data set $\DD=(\DD_1,\ldots,\DD_N)$ that belongs to $\Delta=\{0,1,2\}^N$. Each data item $\DD_k$ is the result of a poll, where a randomly chosen fraction $\varepsilon$ of the $M$ individuals are asked whether they live in a southern or northern city. The result of poll number $k$ is 
$$
\DD_k = \left\{\begin{array}{ll}
2; & \mbox{if $\mathcal S$ is in sample $k$ and $\mathcal S$ answers $\mathsf N$},\\
1; & \mbox{if $\mathcal S$ is in sample $k$ and $\mathcal S$ answers $\mathsf N^c$},\\
0; & \mbox{if $\mathcal S$ is not in sample $k$}.
\end{array}\right.
$$
From this it follows that 
$$
\hmu(\DD) = \left\{\begin{array}{ll}
(d-h)/d; & \DD_{\S}=\emptyset,\\
0; & \DD_{\S}=(1,\ldots,1),\\
1; & \DD_{\S}=(2,\ldots,2).
\end{array}\right.
$$
where $\DD_{\S}=\{\DD_k;\, \DD_k=1\mbox{ or }2\}$ is data for the polls for which $\S$ is among the respondents. Suppose the polls are independent, so that $\DD$ is an observation of a vector $D=(D_1,\ldots,D_N)$ with independent components. Then $D_{\S}=\emptyset$ with probability $(1-\varepsilon)^N$, whereas $D_{\S}=(2,\ldots,2)$ or $D_{\S}=(1,\ldots,1)$ with probability $1-(1-\varepsilon)^N$ depending on whether $x_0\in\mathsf N$ or not. Hence \eqref{muConvLargeN} is satisfied, i.e.\ $\hmu = \P(\TT) \stackrel{p}{\to} \1_{\mathsf N}(x_0)$ as $N\to\infty$, corresponding to full learning asymptotically if $\mathcal S$ tells the truth. From Proposition \ref{Prop:PAsympt}, the limiting posterior distribution $\P_\infty$ of $\P_{\bmu(D)}$ exists a.s. We find that $\P_\infty$ is a uniform distribution on $\mathsf N$ if $x_0\in \mathsf N$, and a uniform distribution on $\mathsf N^c$ if $x_0\notin \mathsf N$. Then, from Definition \ref{D:K}, a necessary condition for $\A$ having full KA asymptotically, as $N\to\infty$, if $\mathcal S$ tells the truth, is $d-h=1$ if $x_0\in \mathsf N$ and $h=1$ if $x_0\notin \mathsf N$. 

Next consider another agent $\tilde{\A}$, whose beliefs differ from those of $\A$ in two ways. Firstly, the prior of $\tilde{\A}$ is based on the assumption that the sizes of the cities $x_k$ of $\EE$ are proportional to $k$. If $\mathcal S$ is a randomly chosen individual from $\EE$, this leads to a prior $\tilde{P}(x_k)=2k/[d(d+1)]\propto k$ for $k=1,\ldots,d$. Secondly, since $\tilde{\A}$ interprets data $\DD$ in a different way than $\A$, he concludes from data that $\mathcal S$ lives in a northern city with probability $\tilde{\mu}=\tilde{\mu}(\DD)$. This may happen, for instance, if $\tilde{\A}$ includes a probability $\delta$ that $\mathcal S$ reports the wrong result in all the polls he takes part in, so that
$$
\tilde{\bmu}(\DD) = \left\{\begin{array}{ll}
(d-h)/d; & \DD_{\S}=\emptyset,\\
\delta; & \DD_{\S}=(1,\ldots,1),\\
1-\delta; & \DD_{\S}=(2,\ldots,2).
\end{array}\right.
$$
From \eqref{Gibbs2}, the posterior beliefs of $\tilde{\A}$ are based on a Gibbs type probability function
\beq
\tilde{P}(x_k) = \left\{\begin{array}{ll}
\frac{2k}{h(h+1)+e^{\tilde{\lambda}}(d-h)(d+h+1)} = \frac{2k(1-\tilde{\mu})}{h(h+1)}; & k=1,\ldots,h,\\
\frac{2ke^{\tilde{\lambda}}}{h(h+1)+e^{\tilde{\lambda}}(d-h)(d+h+1)} = \frac{2k\tilde{\mu}}{(d-h)(d+h+1)}; & k=h+1,\ldots,d.
\end{array}\right.
\lb{tildePSubpop}
\eeq
In terms of Section \ref{Sec:Biased}, we may see $\tilde{\A}$'s beliefs \eqref{tildePSubpop} as a biased version of $\A$'s \eqref{PSubpop}.  
\qed
\end{example}

%\begin{example}[Finite population with several binary features.]\lb{Ex:FiPon} 
%Let us extend the Example \ref{Ex:FiPo1} and consider a finite set \eqref{Xfinite} with $n$ binary features 
%\beq
%	f_i(x) = \1_{\AA_i}(x)
%	\lb{fiFinite}
%\eeq
%that are indicator functions for different subsets $\AA_1,\ldots,\AA_n$ of $\X$. The Gibbs distribution \eqref{Gibbs2} then takes the form
%\beq
%	\P(x_k) = \frac{\exp\left[\sum_{i=1}^n \lambda_i \1_{\AA_i}(x_k)\right]}{\sum_{l=1}^d \exp\left[\sum_{i=1}^n \lambda_i 				\1_{\AA_i}(x_l)\right]}
%	\lb{PFinite}
%\eeq
%for some constants $\lambda_1,\ldots,\lambda_n$ that quantify the impact of each feature on agent $\A$'s posterior beliefs.
%In this case, data in \eqref{D} provide $\A$ with information about the probability $\mu_i = \E_\P f_i(X) = \P(\AA_i)$ of each set $\AA_i$.  
%\end{example}

Example \ref{Ex:FiPo1} motivates Theorem \ref{Prop:FinPop} below. It gives sufficient and necessary conditions for how large $n$ must be to make it possible for $\A$ to attain full KA of any proposition. Therefore, it is a result on the fundamental limits of inference for full KA in classification problems. In what follows, $\lceil x \rceil$ stands for the smallest integer larger or equal to $x$.    

\begin{theorem}[Fundamental limits of knowledge]\lb{Prop:FinPop}
Consider a finite set \eqref{Xfinite} with $d$ elements. Suppose $n$ binary features $f_i(x) = \1_{\AA_i}(x)$
%\beq
%	f_i(x) = \1_{\AA_i}(x)
%	\lb{fiFinite}
%\eeq
are available that are indicator functions for different subsets $\AA_1,\ldots,\AA_n$ of $\X$. %Assume that $\A$'s posterior beliefs \eqref{PFinite} are based on the $n$ binary features \eqref{fiFinite}. 
If $n\ge \lceil \log_2 d \rceil$, it is possible to choose the sets $\AA_1,\ldots,\AA_n$ and constants $\lambda_1,\ldots,\lambda_n$ so that full KA can be attained about any proposition $p$. Conversely, if $n < \lceil \log_2 d \rceil$, for any choice of $n$ binary features, it is possible to pick $x_0$ so that full KA is not possible. 
\end{theorem}

The proofs of Theorems \ref{FLFK} and \ref{Prop:FinPop} are related: The $n$ binary features $f_i(x) = \1_{\AA_i}(x)$ generate a finite partition of $\X$. If $n$ is small, then at least one set of this partition will have more than one element, making full KA impossible for some choices of $p$ and $x_0$.   

\subsection{Coordinatewise features}

In this section, we consider features that are functions of the coordinates of $x$. With two examples, we illustrate that having enough features is crucial for full LKA. 

\begin{example}[One feature per coordinate]\lb{CoordFeat} Assume that 
\beq
	\X = [0,1]^n = \{x=(x_1,\ldots,x_n); 0\le x_i \le 1 \mbox{ for }i=1,\ldots,n\}
\lb{XUnitCube}
\eeq
is the unit cube in $n$ dimensions, with coordinatewise feature functions 
$
	f_i(x) = x_i,  
$
for $i=1,\ldots,n$. We may think of $n$ coins, with $x_0=(x_{01},\ldots,x_{0n})$ containing the probability of heads for each one of them. Data $\DD=(\DD_1,\ldots,\DD_N)\in\Delta=\{0,1\}^{Nn}$ corresponds to flipping the $n$ coins $N$ times, with 
\beq
\DD_k = (\DD_{k1},\ldots,\DD_{kn}) \in \{0,1\}^n
\lb{DDk}
\eeq
the outcome of flip $k$, and with head (tail) corresponding to 1 (0). Assume that $\DD$ is an observation of $D=(D_1,\ldots,D_N)$, with independent components. Assume also that 
\beq
\hbmu = (\hmu_1,\ldots,\hmu_n) = \hbmu(D) = \bar{D} = (\bar{D}_1,\ldots,\bar{D}_n) = \frac{1}{N}\sum_{k=1}^n D_k 
\lb{muDice}
\eeq
is the estimated feature vector of $\A$ containing the fraction of flips for which each coin lands with head. If the prior is uniform on $\X$, $\A$'s beliefs about $x_0$ are given by 
\begin{align}
	P(x)=\prod_{i=1}^n P_i(x_i), \quad
    P_i(x_i) = 
	\left\{\begin{array}{ll}
		1, & \lambda_i = 0,\\
		\frac{\lambda_i e^{\lambda_i x_i}}{e^{\lambda_i}-1}, & \lambda_i\ne 0,
	\end{array}\right. 
\lb{PCoord}
\end{align}
%where
%\beq
%	P_i(x_i) = 
%	\left\{\begin{array}{ll}
%		1, & \lambda_i = 0,\\
%		\frac{\lambda_i e^{\lambda_i x_i}}{e^{\lambda_i}-1}, & \lambda_i\ne 0
%	\end{array}\right. 
%\lb{PiCoord}
%\eeq
and $\lambda_i=\lambda_i(\hmu_i)$. We deduce from \eqref{PCoord} that $\A$'s beliefs about the $n$ coordinates of $x_0$ are independent. However, the discernment $\s$-field is maximal: $\G_\A = \sigma(f_1,\ldots,f_n) =\F$. 

Suppose that $N\to\infty$. By the Law of Large Numbers (LLN), \eqref{muConvLargeN} is satisfied, so %$\bmu(D)$ converges in probability to $\f(x_0)=x_0$ as $N\to\infty$. And from 
$\P_{\hbmu(D)} \stackrel{\mathcal L}{\to} \P_\infty$ a.s., by Proposition \ref{Prop:PAsympt}. Also, Theorem \ref{Thrm:PAsympt} implies that this convergence takes place at rate $1/\sqrt{N}$. It can be seen from \eqref{PCoord} that $\P_\infty$ is different from $\bde_{x_0}$. 
\qed
\end{example}

Theorem \ref{Prop:Coord} shows that full LKA are not warranted for $\A$ in Example \ref{CoordFeat}.

\begin{theorem}\lb{Prop:Coord} 
	In the setting of Example \ref{CoordFeat}, consider propositions $p$ with
	\beq
		\TT = \{x\in [0,1]^n; f_p(x)=1\} = \times_{i=1}^n [a_i,b_i],
		\lb{TTCoord}
	\eeq
	where $0\le a_i < b_i \le 1$ for $i=1,\ldots,n$. If $p$ is %that satisfy \eqref{TTCoord} and are
    true, %($x_0\in\TT$), 
    it is possible for $\A$ to come arbitrarily close to full learning of $p$ 
    if and only if at least one of the two conditions $a_i=0$ and $b_i=1$ holds for each of $i=1,\ldots,n$. Additionally, it is only possible for $\A$ to come arbitrarily close to full KA about $p$ 
    if \emph{all} coordinates of $x_0$ are either 0 or 1.   
\end{theorem}

\begin{example}[Two features per coordinate]\lb{Exa:Coord} Assume $n$ is even and that $\X=[0,1]^{n/2}$ is the unit cube in $n/2$ dimensions. For each coordinate $x_i$, with $i=1,\ldots,n/2$, define one linear and one quadratic feature function
\begin{align*}
f_{2i-1}(x) = x_i, \quad f_{2i}(x) = x_i^2.
\end{align*}
If the prior is uniform on $[0,1]^{n/2}$, then $\A$'s beliefs have density \eqref{PCoord}, with marginals
\beq
P_i(x_i) = \frac{e^{\lambda_{2i-1}x_i + \lambda_{2i} x_i^2}}{\int_0^1 e^{\lambda_{2i-1}t + \lambda_{2i} t^2}\D t}.
\lb{Pxi2FeatCoord}
\eeq
The estimated feature vector $\hmu=(\hmu_1,\ldots,\hmu_n)$ has components $\hmu_{2i-1} = E_\P(X_i)$, $\hmu_{2i}=E_\P(X_i^2)$ for $i=1,\ldots,n/2$ from which it follows that $\Var_\P(X_i)=\hmu_{2i}-\hmu_{2i-1}^2$. In order to describe how $\hbmu$ is generated from data, suppose $x_0=(x_{01},\ldots,x_{0,n/2})$ contains the probability of heads of $n/2$ coins, and that these coins are flipped $N$ times. This gives rise to the same type of data set $\DD=(\DD_1,\ldots,\DD_N)$ as in Example \ref{CoordFeat}, with $\DD_k$ the outcome of flip $k$, defined as in \eqref{DDk} with $n/2$ in place of $n$. Suppose $\DD$ is an observation of $D=(D_1,\ldots,D_N)$, and that the estimated feature vector $\hbmu = \hbmu(D)$ has components 
\begin{align}\lb{muDiceExpVar}
    \hmu_{2i-1} = \bar{D}_i, \quad \hmu_{2i} = \bar{D}_i^2 + \bar{D}_i(1-\bar{D}_i)/N
\end{align}
%\beq
%\begin{array}{rcl}
%\mu_{2i-1} &=& \bar{D}_i,\\
%\mu_{2i} &=& \bar{D}_i^2 + \bar{D}_i(1-\bar{D}_i)/N
%\end{array}
%\lb{muDiceExpVar}
%\eeq
for $i=1,\ldots,n/2$, with $\bar{D}_i$ as defined in \eqref{muDice}. Then, $\bar{D}_i$ and $\bar{D}_i(1-\bar{D}_i)/N$ are the estimated posterior mean and posterior (binomial) variance for the probability of heads of coin $i$. From the LLN, $\hbmu(D)\stackrel{p}{\to} \f(x_0)$, hence Proposition \ref{Prop:PAsympt} implies $\P_{\hbmu(D)}\stackrel{\mathcal L}{\to}\P_\infty$, a.s. This limiting distribution is $\P_\infty = \bde_{x_0}$, as $\bde_{x_0}$ is the limit of a sequence of distributions whose densities $P(x)=\prod_{i=1}^{n/2} P_i(x_i)$ have marginals \eqref{Pxi2FeatCoord}, with $P_i \stackrel{\mathcal L}{\to} \delta_{x_{0i}}$. 
\qed
\end{example}

Example \ref{Exa:Coord} motivates the following result: 

\begin{theorem}\lb{Thrm:Coord}
In the setting of Example \ref{Exa:Coord}, it is possible, by appropriate choice of $\bla$, to come arbitrarily close to full learning and full KA of any proposition $p$ such that %either a) $p$ is true and $x_0$ is an interior point of the truth set $\TT$ in \eqref{TT}, or b) $p$ is false and $x_0$ is an interior point of $\TT^c$.  
$p$ is true (false) and $x_0$ is an interior point of $\TT$ ($\TT^c$).    
\end{theorem}

Theorem \ref{Thrm:Coord} shows that two features per coordinate of $x$ make it possible for agent $\A$ to acquire feature data $\DD$, with $\bla=\bla(\hbmu(\DD))$ chosen so that he gets arbitrarily close to full LKA of proposition $p$. In contrast, Theorem \ref{Prop:Coord} reveals that it is typically not possible for $\A$ to get close to full LKA of $p$ when only one feature per coordinate of $x$ is available (regardless of the size $N$ of the dataset $\DD$). With one feature per coordinate, $\A$ can only vary the expected value $\hmu_i$ of his beliefs about each coordinate $x_i$ of $x$. With two features per coordinate, $\A$ is able to vary the expected value $\hmu_{2i-1}$ {\it and} the variance $\hmu_{2i}-\hmu_{2i-1}^2$ of his beliefs about $x_i$. Theorem \ref{Thrm:Coord} refers to the limit when $\hmu_{2i-1}$ converges to $x_{0i}$ (component $i$ of $x_0$) and $\hmu_{2i}-\hmu_{2i-1}^2$ converges to 0, in agreement with \eqref{muDiceExpVar}. 

\begin{remark}
Examples \ref{CoordFeat} and \ref{Exa:Coord} can be generalized to the case when $\X=[0,1]^{n/m}$ and there are $n$ feature functions 
$
f_{mi-m+j}(x) = h_j(x_i)
$,
obtained from $m$ basis functions $h_1,\ldots,h_m$ for each coordinate $i=1,\ldots,n/m$. An option is to use polynomials $h_j(x_i)=x_i^j$. Another option is to choose $\{h_j\}_{j=1}^m$ as kernel functions from a reproducing Hilbert space \cite{GrettonEtAl2012, ScholkopfSmola2002, SriperumbudurEtAl2010}. We conjecture that the latter choice of basis functions can be very useful for the $n$-dimensional space of Gibbs distributions \eqref{Gibbs2} to accurately approximate the space of probability distributions on $\X$ with independent marginals. These basis functions can also be efficiently computed from a random feature map \cite{RahimiRecht2007}.    
\qed    
\end{remark}

\subsection{Piecewise constant posterior}\lb{Sec:Piecewise}

We present two examples with features that lead to piecewise constant posteriors $\P$. For this class of features, full KA is not possible, although full learning sometimes is.  

\begin{example}[Piecewise constant posterior in one dimension.]\lb{Ex:PieceConst}
	Suppose $\X=[0,1)$ is the half-open unit interval, which is divided into $n$ equally large and disjoint sets $\AA_i = [(i-1)/n,i/n)$ for $i=1,\ldots,n$. The feature functions $f_i(x) = \1_{\AA_i}(x)$
	%\beq
	%	f_i(x) = \1_{\AA_i}(x)
	%	\lb{fiPiecewise}
	%\eeq
are indicator functions for these intervals. Suppose $x_0$ is the probability of heads of a coin. Data $\DD=(\DD_1,\ldots,\DD_N)\in \Delta=\{0,1\}^N$ is the outcome of flipping this coin $N$ times, with 1 (0) corresponding to heads (tails) in each flip. Assume that $\DD$ is an observation of $D=(D_1,\ldots,D_N)$, and let $\bar{D} = (D_1 + \cdots + D_N)/N$
%\beq
%\bar{D} = \frac{1}{N}\sum_{k=1}^{\textcolor{red}{N}} D_k
%\lb{barD}
%\eeq
be the fraction of heads from the $N$ flips. This gives rise to estimated features 	
$
\hmu_i= \hmu_i(D) = \E_\P f_i(X) = \P(\AA_i) = \1\{\bar{D}\in \AA_i\}
$ 
for $i=1,\ldots,n$. Assume also that the ignorant agent $\I$ has a uniform density $P_0(x)=1$ on $\X$. %, according to \eqref{P0Unif}. 
Then, $\A$'s posterior density \eqref{Gibbs2} is piecewise constant 
	\beq
		P(x) = \sum_{i=1}^n p_i \1_{\AA_i}(x)
	\lb{PPiecewise}
        \eeq
	over each $\AA_i$, as in \eqref{PPiecewise2}, with values  
	\beq
		p_i = n\hmu_i = ne^{\lambda_i}/\left(e^{\lambda_1}+\ldots+e^{\lambda_n}\right) \propto e^{\lambda_i}.
	\lb{pi}
        \eeq

Note that the feature functions are linearly dependent: $f_1(x) + \cdots f_n(x) = 1$.
%\beq
%\sum_{i=1}^n f_i(x) = 1.
%\lb{fiLD}
%\eeq
For this reason, one of them is redundant. Nonetheless, it is still convenient to have $n$ (rather than $n-1$) feature functions because of symmetry. This linear dependency implies, however, that $\bla$ does not uniquely characterize $\P$ since we may add the same constant to all $\lambda_i$ without changing $\P$. Without loss of generality, we can therefore assume that $\bla$ is chosen so that the last proportionality of \eqref{pi} is an equality, which implies that $n = e^{\lambda_1} + \cdots + e^{\lambda_n}$. We conclude %from \eqref{GAGibbs} and \eqref{fiPiecewise} 
that
$
		\G_\A = \sigma(\AA_1,\ldots,\AA_n) 
$	
is the set of all $2^n$ finite unions of sets $\AA_i$. %(this corresponds to a finite partition of $\X$ of size $n$ in \eqref{GPartition2}, in order to generate $\G_\A$). 
Hence, $1/n$ is the maximal resolution by which $\A$ is able to discern between different possible worlds. This can also be seen by letting the size $N$ of the dataset increase: from %\eqref{barD} and 
the LLN, $\bar{D}\stackrel{p}{\to}x_0$, as $N\to\infty$. Then 
\beq
\hbmu_i(D)\stackrel{p}{\to} \1\{i=i_0\}
\lb{bmuiDConv}
\eeq
for $i=1,\ldots,n$, if $x_0$ is an interior point of $\AA_{i_0}$. From \eqref{PPiecewise}, \eqref{pi}, and \eqref{bmuiDConv} we deduce, by Proposition \ref{Prop:PAsympt}, that the posterior of agent $\A$ converges to a uniform distribution, 
\beq
\P\stackrel{\mathcal L}{\to}\P_\infty = \mbox{Unif}(\AA_{i_0})
\lb{PConvUnif}
\eeq
as $N\to\infty$, a.s. However, when $x_0=(i_0-1)/n$ is at the boundary between $\AA_{i_0-1}$ and $\AA_{i_0}$, it follows from the Central Limit Theorem applied to $\sqrt{N}(\bar{D}-x_0)$ that \eqref{PConvUnif} does not hold. Instead, when $N$ gets large, $\P$ equals either $\mbox{Unif}(\AA_{i_0-1})$ or $\mbox{Unif}(\AA_{i_0})$ with equal probabilities 0.5. 

Suppose $n=10$. Then $\AA_i\subset\X$ consists of all $x$ whose first decimal is $i-1$, and the posterior \eqref{PPiecewise} corresponds to $\A$:s beliefs about the first decimal of $x_0$. The proposition 
	$$
		p: \mbox{The first decimal of $x_0$ is 5}
	$$
has truth function $f_{p}=f_{6}$, and the set of worlds for which $p$ is true is $\TT=\AA_{6}$. It follows from \eqref{PConvUnif} and the paragraph below, that $\A$ will (will not) fully learn $p$ as $N\to\infty$ when $x_0\notin \{0.5,0.6\}$ (when $x_0\in\{0.5,0.6\}$). But in the former case, since $\A$ only knows whether $x_0\in \AA_{6}$ asymptotically, he still does not attain full KA of $p$ asymptotically. Indeed, suppose for instance $p$ is true and $\hmu_{6}=1$. It follows then from \eqref{PPiecewise}, that for any $\varepsilon < 1/(2n)=1/20$ the posterior probability of the open ball $\BB = B_\epsilon(x_0)$ is 
    \beq
    \P(\BB) = 1 - \P(\BB^c) = 1 - n |\AA_{6}\setminus \BB| \le 1 - n(\frac{1}{2n}-\varepsilon) = \frac{1}{2} + n\varepsilon < 1,
    \lb{PPiecewiseLower}
    \eeq
independently of $N$. Consider now a second proposition
    $$
    p^\prime: \mbox{The second decimal of }x_0\mbox{ is 5}, 
    $$
    with $\TT^\prime$ the set of worlds for which $p^\prime$ is true. Since $n=10$, it is clear that 
    $\P(\TT^\prime) = \P_0(\TT^\prime) = 0.1$,
    regardless of the choice of $\P$ in \eqref{PPiecewise}. Hence, $\A$ does not learn anything about $p^\prime$ (the second decimal of $x_0$), no matter how accurate information he receives about the first decimal of $x_0$. This is an illustration of Theorem \ref{FLFK}, where it is not only impossible for $\A$ to learn $p^\prime$ fully, but it is not even possible for $\A$ to learn anything at all about $p^\prime$. In order for $\A$ to learn about $p^\prime$, he needs to add features about the second decimal of $x$, corresponding to $n=100$. This makes it possible for $\A$ to fully learn $p^\prime$ (when $x_0$ is not a boundary point of $\TT^\prime$), although he still does not acquire full knowledge about $p^\prime$ (cf.\ \eqref{PPiecewiseLower}). 
\qed
\end{example}

Next, we generalize Example \ref{Ex:PieceConst} by considering an $r$-dimensional piecewise constant posterior, obtained from a recursively partitioned binary tree, which is significant because this structure is used to construct classification and regression trees \cite{BreimanEtAl1984, Ripley1996}. The details of its construction %of the recursively partitioned binary tree 
and the corresponding posterior distribution $\P$ are given in the proof in the Supplement \cite{DiazEtAlSupp2025}. 

\begin{theorem}\lb{Th:RT}
    Let $\X=[0,1]^r$ and let $\PP = \{\AA_1,\ldots,\AA_n\}$ be a finite partition of $\X$ that is obtained as a recursively partitioned binary tree, so that all $\AA_i$ are rectangles with sides parallel to the coordinate axes. Then, full KA is only attained in the limit when the number of features $n$ goes to infinity.
\end{theorem}

\subsection{A mixture of a continuous and discrete posterior}\lb{Sec:LimDisc}

Example 1 of the Supplement \cite{DiazEtAlSupp2025} presented a $\sigma$-field that turned out to be inappropriate for representing $\A$'s discernment, since Definition \ref{DefDis} is violated. Here we will approximate this $\sigma$-field with a smaller one $\G_\A$, whose resolution requires the posterior distribution of agent $\A$ to be a mixture of a continuous and a discrete distribution. Since this distribution is not a Gibbs distribution \eqref{Gibbs2}, we will in turn approximate $\G_\A$ with another $\s$-field $\tilde{\G}_{\tilde{\A}}$ that gives rise to posteriors that are Gibbs distributions, with piecewise constant densities, as in Example \ref{Ex:PieceConst}. Although this represents an information loss, this loss can be made arbitrarily small by decreasing the lengths of the intervals along which the posterior is constant. This is all contained in the following proposition.%, which is proved in the Supplement \cite{DiazEtAlSupp2025}: 

\begin{proposition}\lb{CounterBill} \ 
    \begin{enumerate}
    \item Let $\AA = \{x_1,x_2,\ldots\} \subset [0,1]$ be a fixed countable set, and define the $\s$-field
        $
            \G_\A = \s([0,1]\setminus \AA,x_1,x_2,\ldots),
        $
        generated by the complement of $\AA$ and the elements of $\AA$ (or equivalently, the collection of sets $\BB$ such that either $\BB$ or $\BB^c$ is a subset of $\AA$). Even though it is not possible to express the posterior as a Gibbs distribution, it is sometimes possible to fully learn and acquire full knowledge about a proposition $p$ with truth set $\TT$. Full learning is possible if either $p$ is true and $\AA\cap\TT\ne \emptyset$ or if $p$ is false and $\AA\cap\TT^c \ne \emptyset$. Full KA can be attained if, additionally, $p$ is true and $x_0\in\AA\cap\TT$, or if $p$ is false and $x_0\in \AA\cap\TT^c$. 
    \item Let 
        $
            \tilde{\G}_\A = \s([0,1]\setminus \tilde{\AA},x_1,x_2,\ldots,x_n)
        $
        be obtained from the finite set $\tilde{\AA}=\{x_1,\ldots,x_n\}$. Then, it is possible to approximate the posterior with a Gibbs distribution of $n$ features. Full learning is possible under the same conditions as in Part 1, with $\tilde{\AA}$ in place of $\AA$. KA is possible under the same conditions, to a degree that depends on how well the Gibbs distribution approximates the posterior.     
    \end{enumerate}
\end{proposition}

\section{Secondary learning and knowledge acquisition}\lb{Sec:LKAGibbsSec}

In this section, we analyze secondary learning, whereby an agent $\tilde{\A}$ learns about the learning of another agent $\A$. Recall that agent $\A$ has primary data $\DD$ from some space $\Delta$, from which he infers estimates of the values of $n$ features. This makes it possible for him to form beliefs about $x\in\X$ according to the Gibbs posterior density $P(\cdot;\bla)$ in \eqref{Gibbs2}, where $\bla=\bla(\hbmu(\DD))$. Agent $\tilde{A}$, on the other hand, has secondary data $\tilde{\DD}$ from some other space $\tilde{\Delta}$ that makes it possible for him to learn about $\A$'s learning. This is to say that $\tilde{\A}$ learns about the Gibbs posterior density $P(\cdot;\bla)$ of $\A$. Note in particular that the interpretation of $\bla$ differs between $\A$ and $\tilde{\A}$. For agent $\A$, $\bla=\bla(\hbmu(\DD))$ is a sufficient statistic for doing inference about the parameter $x$, based on the data $\DD$ that he receives. On the other hand, for agent $\tilde{A}$, $\bla$ is a parameter of $\A$'s posterior beliefs that needs to be estimated as part of his learning about $\A$'s learning. Therefore, the secondary data $\tilde{\DD}$ of $\tilde{\A}$ should provide information about $\bla$ (and only indirectly about $x$). In more detail, we will assume that $\tilde{\A}$ receives a random sample $\tilde{\DD} = \{x_1,\ldots,x_m\}$
%\beq
%\tilde{\DD} = \{x_1,\ldots,x_m\}
%\lb{tD}
%\eeq
of size $m$ from $\A$'s parameter space $\X$, so that $\tilde{\Delta}=\X^m$. We will consider two scenarios, where agent $\tilde{\A}$ either forms his beliefs about $x_0$ using a maximum likelihood approach (Section \ref{Sec:ML}) or a Bayesian approach (Section \ref{Sec:Bayes}) in order to estimate $\bla$. As a preparation, in Section \ref{Sec:OptEmp} we will first introduce optimization (maximum likelihood estimation of $\bla$) under empirical (secondary type of learning) side constraints.

\subsection{Optimization under empirical side constraints}\lb{Sec:OptEmp} 

A variant of the optimization problem (\ref{Popt})-(\ref{muiQP}) is to assume that features are estimated from a sample $\tilde{\DD}=\{x_j\}_{j=1}^m$ from $\X$. This corresponds to replacing  (\ref{muiQP}) with constraints 
\beq
\mu_i(\bpi) = \frac{1}{m}\sum_{j=1}^m f_i(x_j), \quad i=1,\ldots,n,
\lb{muiQpi}
\eeq
where $\bpi=\sum_{j=1}^m \bde_{x_j}/m$ is the empirical distribution corresponding to $\tilde{\DD}$, whereas $\bde_x$ refers to a point mass at $x$. It has been shown in \cite{DellaPietra2Lafferty1997} that the solution to the maximization problem \eqref{Popt} is given by density function $\tilde{P}(x) = Q_{\hat{\boldsymbol{\lambda}}}(x)$, where 
\begin{align}\label{hla}
		\hat{\boldsymbol{\lambda}} &= \mbox{arg} \max_{\boldsymbol\lambda \in \R^n} \prod_{j=1}^m 								Q_{\boldsymbol{\lambda}}(x_j)
				 = \mbox{arg}\max_{\boldsymbol\lambda \in \R^n} \prod_{x \in \X} Q_{\boldsymbol{\lambda}}(x)^{m \pi(x)} 
				= \mbox{arg}\max_{\boldsymbol\lambda \in \R^n} \sum_{x \in \X} \pi(x) \log Q_{\boldsymbol{\lambda}}(x)\nonumber \\
				 &= \mbox{arg} \max_{\boldsymbol\lambda \in \R^n} \E_{\bpi} [\log Q_{\boldsymbol{\lambda}}(X)] 
				= \mbox{arg} \min _{\boldsymbol\lambda \in \R^n} 	\infdiv{\bpi}{\QQ_{\boldsymbol{\lambda}}}
\end{align}
is the maximum likelihood estimator of $\boldsymbol{\lambda}$, when $\tilde{\DD}$ is viewed as a sample of iid observations from the Gibbs distribution (\ref{Gibbs2}). From the third step of (\ref{hla}) we find that $\QQ_{\hat{\boldsymbol{\lambda}}}$ is the Gibbs distribution that maximizes the cross entropy between $\bpi$ and $\QQ_{\boldsymbol{\lambda}}$. That is, $\QQ_{\hat{\boldsymbol{\lambda}}}$ minimizes the expected log loss $\E_{\bpi}[-\log Q_{\boldsymbol{\lambda}}(X)]$ among all Gibbs distributions. It has further been noted (see, e.g., \cite{BoydVandenbergue2004, CelisKeswaniVishnoi2020, DudikPhillipsSchapire2007}) that the following are convex optimization programs equivalent to those in \eqref{hla}:
\begin{align}\label{equivalences3}
\begin{aligned}
\hat{\bla} = \mbox{arg}\max_{\boldsymbol\lambda \in \R^n} \E_{\bpi} \left( \log [Q_{\boldsymbol{\lambda}}(X)/P_0(X)] \right) 
= \mbox{arg}\max _{\boldsymbol\lambda \in \R^n} 	[\infdiv{\bpi}{\P_0} -  \infdiv{\bpi}{\QQ_{\boldsymbol{\lambda}}}].
\end{aligned}
\end{align}
In particular, from the second step of (\ref{equivalences3}) we deduce that $\hat{\bla}$ maximizes the expected value $\E_{\bpi}[I^+(\{X\};\P_0,\QQ_{\bla})]$ of an AIN measure.

\subsection{Maximum likelihood plug-in approach to secondary learning}\lb{Sec:ML}

In this section we assume that $\tA$ forms his beliefs about $\A$'s beliefs about $x_0$, from the plug-in posterior density 
\beq
\tilde{P}(x) = P(x;\hbla) = Q_{\hbla}(x),
\lb{tP}
\eeq
where $\hbla$ is the maximum likelihood estimator of $\bla$, defined in \eqref{hla}. It follows from (\ref{AINA}) and (\ref{AINA2}) that agent $\tA$ believes that $\A$ has learnt an amount
\beq
    \hat{I}^+(\TT) = I^+(\TT) + \mbox{Bias} ( \TT;\bla,\hbla)
    \lb{I+tP}
\eeq
about $p$, where $I^+(\TT)=I^+(\TT;\P_0,\P)$ is the actual amount of learning of $\A$ about $p$, whereas $\hat{I}^+(\TT)=I^+ \left(\TT;\P_0,\tP \right)$ is $\tA$'s estimate of this quantity. The following proposition gives an asymptotic expansion of $\tA$'s expected estimate of $\A$'s learning: 

\begin{proposition}\lb{Prop:SecLearning}
Suppose $\tA$ forms his beliefs about $\A$'s beliefs in $x_0$ according to \eqref{tP}, based on a secondary learning data set $\tilde{\DD}$ of size $m$, an observation of a random sample $\tilde{D}$ with independent components drawn from $\A$'s posterior distribution $\P=\QQ_{\bla}$ in \eqref{Gibbs2}, with $\bla=\bla(\hbmu(\DD))$ obtained from $\A$'s primary learning dataset $\DD$. Then asymptotically, $\tA$'s expected secondary learning about $\A$'s beliefs in proposition $p$ is    
\beq
\E [\hat{I}^+(\TT)] = I^+(\TT) + \frac{C}{m} + o \left(m^{-1}\right)
\lb{EI+tp}
\eeq
as $m\to\infty$, where $\TT$ is the set of worlds \eqref{TT} where $p$ is true, and expectation is taken wrt random variations in $\tilde{D}$. Moreover, $C = \mbox{tr}\left(\J^{-1}\H\right)/2$, $\J = \J(\bla) = \E_{\QQ_{\bla}} \left[\f(X)\f(X)^T\right]$ is the Fisher information matrix that corresponds to the maximum likelihood estimate \eqref{hla} of $\bla$, and $\H$ is the Hessian matrix of the function $\bla^\prime \to \mbox{Bias}\left(\TT;\bla,\bla^\prime\right)$ at $\bla^\prime=\bla$.%, whereas $o\left(m^{-1}\right)$ is a remainder term that is small in comparison to $m^{-1}$ as $m\to\infty$.
\end{proposition}

\subsection{Bayesian approach to secondary learning}\lb{Sec:Bayes}

Has $\tA$ learned and acquired knowledge about $p$? Not necessarily, since $\tA$ tries to recapitulate the beliefs of $\A$ about $p$, based on data $\tilde{\DD}$, without having access to original data $D$ that $\A$ used in order to formulate his beliefs about $p$. Since $\tilde{\A}$ does not take the trouble to process original data to form his beliefs, it is safer to say that $\tilde{\A}$ learns and acquires knowledge about how much $\A$ has learned about $p$. 
This corresponds to an LKA problem with a true world 
$$
\tilde{x}_0=I^+(\TT)\in \left(-\infty,-\log \P_0(\TT)\right] =: \tilde{\X}.
$$
In order to define this LKA problem properly, in line with Section \ref{Sec:LKA}, in this section we take a Bayesian approach about $\bla$ and treat it as a random parameter with a prior density $P_0(\bla)$ and posterior density 
\beq
	\tilde{P}(\bla)\propto \tilde{L} \giventhat*{\tilde{\DD}}{\bla} P_0(\bla), 
	\lb{tildePla}
\eeq 
where $\tilde{L} \giventhat*{\tilde{\DD}}{\bla}$ is the likelihood defined in the first line of \eqref{hla}, used by agent $\tilde{\A}$ in order to make inference about $\bla$. This gives rise to a modified version  
\beq
\tilde{P}(x) = \int P(x;\bla)\tilde{P}(\bla)\D\bla
\lb{tP2}
\eeq
of (\ref{tP}), that is, a modified version of agent $\tilde{\A}$'s expected beliefs about $\A$'s beliefs about $x_0\in\X$. In order to formalize $\tilde{\A}$'s learning about $\A$'s learning, consider the proposition 
$$
\tilde{p}: \mbox{Agent }\A \mbox{ has increased his beliefs that }p\mbox{ is true}.
$$
This proposition is true if $\tilde{x}_0 = I^+(\TT)\in \left(0,-\log \P_0(\TT)\right]:=\tilde{\TT}\subset\tilde{\X}$. Hence, agent $\tilde{\A}$'s learning about $\tilde{p}$ is given by 
$
    \tilde{I}^+ \left(\tilde{\TT}\right) = \log \tilde{\P} \left(\tilde{\TT}\right) -\log \P_0\left(\tilde{\TT}\right),
$
where 
\begin{align*}
		\P_0 \left(\tilde{\TT}\right) = \int \1 \left[I^+(\TT;\bla)>0\right] P_0(\bla)\D\bla, && \mathrm{and} &&
		\tilde{\P} \left(\tilde{\TT}\right) = \int \1\left[ I^+(\TT;\bla)>0\right] \tilde{P}(\bla)\D\bla
\end{align*}
represent agent $\tilde{\A}$'s beliefs in $\tilde{\TT}$ before and after he received data $\tilde{\DD}$ respectively, where the RHS of the last two equations use the simplified notation $I^+(\TT;\bla)=I^+(\TT;\P_0,\QQ_{\bla})$. In addition, $\tilde{\A}$ also learns and acquires knowledge about how much knowledge $\A$ has acquired about $p$. This corresponds to a LKA problem with a true world $\tilde{x}_0 = P(\cdot;\bla)\in \QQQ =: \tilde{\X}$, where $\QQQ$ is the set of distributions on $\X$. From the posterior distribution (\ref{tildePla}) of $\bla$ given data $\tilde{\DD}$, it is possible to compute a posterior distribution of the density $P(\cdot;\bla)$ given data $\tilde{\DD}$ for agent $\tilde{\A}$. The latter posterior distribution can be used to define various aspects of agent $\tilde{\A}$'s LKA about $\A$'s KA about $p$. 

\section{Discussion}\lb{Sec:Disc}

\subsection{Summary}

In this paper, we have used the concept of AIN to analyze LKA of a proposition $p$ for an agent $\A$ who receives primary data $\DD$ in terms of a number of features of relevance for $p$. This leads to a Gibbs distribution for the posterior distribution that corresponds to the beliefs of $\A$ about the true explanation $x_0$ of $p$. We also introduced the concept of secondary learning for an agent $\tilde{\A}$ who does not have access to original data $\DD$ but rather receives data $\tilde{\DD}$ from $\A$. Our work has implications for statistical learning, where an algorithm $\A$ receives data on a number of features of an object $x_0$ in order to learn and acquire knowledge about various propositions of relevance for the object. We have highlighted potential limitations of such statistical learning algorithms based on feature extraction: When the number of features is too small, this type of primary learning is not always possible, and full KA is not guaranteed. This in turns sets limits on $\tA$:s secondary learning.     

\subsection{Extensions}

%The results of this article can be extended in various ways, as described in the sections below.  

\subsubsection{6.2.1 The dynamics of primary and secondary learning}

One can look at LKA dynamically as a function of the size of the data set. This holds for primary data $\DD=(\DD_1,\ldots,\DD_N)$ as well as for secondary data $\tilde{\DD}=(x_1,\ldots,x_m)$. Recall that these datasets are observations of random vectors $D=(D_1,\ldots,D_N)$ and $\tilde{D}=(X_1,\ldots,X_m)$ respectively. Hence we can view the dynamics of primary and secondary LKA from a stochastic process point of view, as a function of $N$ and $m$ respectively. For primary data, Proposition \ref{Prop:PAsympt} gives conditions under which agent $\A$'s beliefs $\P=\P_N$ converge towards a limiting posterior distribution $\P_\infty$. For secondary learning, agent $\tilde{\A}$'s posterior distribution $\tilde{\P}=\tilde{\P}_m$ converges to $\P$ as $m\to\infty$. The components $X_j$ of $\tilde{D}$ need not be observations of independent random variables with distribution $\P$, but more generally $\tilde{D}$ could be a Markov process with stationary distribution $\P$. Under certain conditions the resulting learning process could be described through Glauber dynamics or Metropolis-Hastings algorithms \cite{LevinPeres2017}. This makes it possible to analyze various asymptotic properties of the secondary learning process.

\subsubsection{6.2.2 Asymptotic knowledge acquisition}

An important aspect of the dynamics of KA (Section 6.2.1) is whether the asymptotic posterior distribution equals a point mass $\bde_{x_0}$ at $x_0$ and thereby corresponds to full KA about $x_0$. This is typically not the case for secondary learning, since the asymptotic limit of agent $\tilde{\A}$' s posterior $\tilde{\P}$ is $\P$ rather than $\bde_{x_0}$. For primary learning, the asymptotic limit $\P_\infty$ of agent $\A$'s posterior 
will depend on $n$, the number of features of data. Since the number of features sets a limit to the resolution of $\A$'s posterior beliefs, it follows that $N\to\infty$ is not a sufficient condition for having full KA asymptotically. In the present article, we used a combined method of moments and quasi-Bayesian approach to find the posterior distribution of agent $\A$. Since this posterior distribution is not based on a true likelihood, traditional Bayesian asymptotic theory is not directly applicable to finding the asymptotic limit $\P_\infty$ of $\P$ as $N\to\infty$. Note however that Proposition \ref{Prop:PAsympt} implies asymptotic full KA as $N\to\infty$, when the number of features is large enough to warrant a limiting posterior $\P_\infty=\bde_{x_0}$.  

In contrast, in \cite{HossjerDiazRao2022} we used a proper likelihood to define the posterior beliefs of $\A$ through Bayes Theorem. As long as the true world $x_0$ is identifiable from the likelihood, the model is correctly specified and there is sufficient prior mass around $x_0$, a posterior distribution based on a true likelihood will asymptotically be concentrated at $x_0$. According to \cite{GhosalvanderVaart2017}, this can be formalized through the following two conditions: Firstly, the posterior distribution $\P$ converges at rate $\epsilon_N\to 0$ towards $x_0$ if 
\beq
\lim_{N\to\infty} \E_{x_0}[\P(B_{M_N\epsilon_N}(x_0)^c)] = 0 
\lb{AsP}
\eeq
for all sequences $M_N\to\infty$, with $B_{\epsilon}(x_0)$ the open ball \eqref{Bepsx0} of radius $\epsilon$ around $x_0$, and with $\E_{x_0}$ referring to expectation of data $D$ when $x_0$ is the true parameter. Secondly, to ensure that $x_0$ is asymptotically included in the support of $\P$, let $\hat{\AA}_N$ be a credibility set, with a level of confidence $0<\alpha < 1$, computed from a posterior $\P=\P_{\hbmu(D_1,\ldots,D_N)}$ based on $N$ data items. Then, the second condition for asymptotic convergence is
\beq
\liminf_{N\to\infty} \P_{x_0}(x_0\in \hat{\AA}_N) \ge 1-\alpha,
\lb{AsCov}
\eeq
with $\P_{x_0}$ referring to probabilities for data $D=(D_1,\ldots,D_N)$ when $x_0$ is the true parameter. Section 7.2 of \cite{HossjerDiazRao2022} was devoted to Bayesian asymptotic theory. In particular, in \cite[Remark 11]{HossjerDiazRao2022}, we made a comment that \eqref{AsP} is equivalent to having full KA asymptotically at rate $\epsilon_N$. In \cite{HossjerDiazRao2022} we also considered the special case where $\X$ is a subset of Euclidean space and the components $D_k$ of $D$ are iid. Bernstein--von Mises Theorem and asymptotic normality of maximum likelihood estimators were used to conclude that the posterior $\P$ is approximated by a Gaussian distribution with covariance matrix of order $N^{-1}$, with a mode whose distance to $x_0$ is also normally distributed with the same asymptotic covariance matrix. It can be shown that this implies that \eqref{AsP} holds with $\epsilon_N=N^{-1/2}$, whereas \eqref{AsCov} holds for all $0<\alpha<1$. We conjecture that results analogous to \eqref{AsP} and \eqref{AsCov} can also be established in the quasi-Bayesian context of the present article, when the number features $n$ is large enough to warrant $\P_\infty=\bde_{x_0}$. For instance, when the components of $D=(D_1,\ldots,D_N)$ are independent, it follows, under the conditions of Theorem \ref{Thrm:PAsympt}, that \eqref{AsP} holds with $\epsilon=N^{-1/2}$ when $\P_\infty=\bde_{x_0}$.  

\subsubsection{6.2.3 Synthetic primary learning versus secondary learning for language models}

There are other types of artificial data sets than secondary data $\tilde{\DD}$ that can be used for LKA. One such example is synthetic primary data $\DD^\prime$ produced, for instance, by large language models (LLMs). It is possible that one of the reasons why LLMs sometimes produce outputs with high error rates (such as confidently hallucinating non-existing facts, using outdated knowledge, generating non-transparent reasoning or toxic outputs that may offend or discriminate) is that they are trained on synthetic data generated by other LLMs, see \cite{CherianGibbsCandes2024, HuangEtAl2025} and references therein. It has been found in \cite{LiEtAl2023} that the performance of LLMs that are trained on synthetic primary data is worse for tasks with high subjectivity (such as humor and sarcasm detection) than for tasks with low subjectivity (such as news topics classification and email spam detection). To illustrate synthetic primary learning versus secondary learning in the context of humor detection, suppose a query is made whether a given sentence $S$ is humorous or not. This can be formulated as a proposition (or claim) $p$ that $S$ is humorous, whereas the true world $x_0\in\X$ is the reason why $S$ is humorous (if $x_0\in\TT$) or not (if $x_0\notin\TT$). Primary test data $\DD\in\Delta$ consist of $N$ sentences generated by humans that are tagged as humorous or not, on which an LLM $\A$ is trained. In order to analyze data, $\A$ makes use of $n$ complementary rules (or features) to determine whether a sentence is humorous or not. Note that primary data goes beyond using $\A$'s internal knowledge from large language models, in that it also makes use of external knowledge bases (for instance through Retrieval-Augmented Generation, \cite{GaoEtAl2023b}). Primary synthetic test data $\DD^\prime\in\Delta$, on the other hand, does not include external knowledge. It consist of sentences generated by $\A$ that are tagged as humorous or not, on which another LLM $\A^\prime$ is trained, making use of the same $n$ rules. In contrast, secondary data $\tilde{\DD}$ consist of $m$ (correct or incorrect) tentative explanations of $\A$, as to why $S$ is humorous or not. This data $\tilde{\DD}$ could be used by a human $\tilde{\A}$ who consults $\A$ to determine whether $S$ is humorous or not, and why.

In other contexts, we may think of the secondary agent $\tilde{\A}$ as an LLM who answers a query by searching a large database for answers to the query. Suppose a sample $\tilde{\DD}=(x_1,\ldots,x_m)$ of putative answers to the query are found, and that the database contains texts from a large number $L$ of humans. We may then think of $\tilde{\DD}$ as the output from an agent $\A$ that represents all $L$ individuals that contributed with data. The posterior $\P=\sum_{l=1}^L w_l \P_l$ of $\A$ is a weighted average ($w_l\ge 0$, $\sum_l w_l=1$), with $\P_l$ and $w_l$ the posterior beliefs and fraction of data in the database, for individual $l$. 

Synthetic primary learning can be modeled mathematically as follows: Recall that primary data $\DD$ is used by agent $\A$ to make inferences about $x_0$. This primary data is an observation of a random variable $D$ on $\Delta$, whose distribution is assumed to follow the mixed likelihood $\int L(\cdot|x_0)P_0(x)\D x$ of agent $\I$ (although the true likelihood, for data generated without bias, is $L(\cdot|x_0)$). Recall also that secondary data $\tilde{\DD}\in \tilde{\Delta}=\X^m$ is an independent sample of size $m$, generated by agent $\A$ from the distribution $\P$ on $\X$ that constitutes his beliefs about $x_0$. Synthetic primary data, on the other hand, is artificial primary data generated by $\A$. It can be viewed as an observation of a random variable $D^\prime$ on $\Delta$ whose distribution follows the mixed likelihood $L(\cdot) = \int L(\cdot|x)P(x)\D x$ of $\A$. Consequently, $\DD^\prime$ and $\tilde{\DD}$ are both generated by agent $\A$, but for the different purposes of producing new (artificial) primary data and informing about the beliefs of $\A$ respectively. In spite of this, synthetic primary data will have similar asymptotic consequences as secondary data. To motivate this, assume that synthetic primary data $\DD^\prime = (\DD^\prime_1,\ldots,\DD^\prime_{N^\prime})$ of size $N^\prime$ is available to agent $\A^\prime$, whose components are observations of independent and identically distributed random variables in $D^\prime=(D_1^\prime,\ldots,D_{N^\prime}^\prime)$. 
Analogously to Proposition \ref{Prop:PAsympt}, if we let $N^\prime\to\infty$, it then follows that $\hbmu(D^\prime)\stackrel{p}{\to}\bmu(\P)$, and consequently $\P^\prime\stackrel{\mathcal L}{\to}\P$, since $\P$ is the Gibbs distribution that corresponds to the limiting observed feature vector $\bmu(\P)$ of $\A^\prime$. This is to say that the posterior distribution $\P^\prime$ of agent $\A^\prime$ (just as the posterior distribution $\tilde{\P}$ for agent $\tilde{\A}$) converges to $\P$ rather than to a point mass at $x_0$, as the size of the data set increases.  

The conclusion is that neither synthetic primary data nor secondary data will generate full knowledge asymptotically about a proposition as the size of data grows, unless agent $\A$ has already acquired full knowledge about this proposition. In the context of humor detection, agents $\A^\prime$ and $\tilde{\A}$ will never learn beyond $\A$'s interpretation on whether sentence $S$ is humorous or not, and they will never be able to explain why $S$ is humorous or not, beyond the explanations provided by $\A$. More generally, for propositions $p$ that either concern rare events and/or relate to moral, ethical, and religious issues, it seems that synthetic primary learning and secondary learning algorithms are subject to bias, since these two types of learning ultimately depend on others learning about $p$ rather than on primary data of relevance for $p$. These observations reinforce our claim in Section \ref{Sec:LKAIntro} that statistical learning does not always entail knowledge. 

\subsubsection{6.2.4 Learning and fine-tuning}

The results in this article have implications for learning whether a particular object $x_0$ from a set $\X$ of possible objects is finely tuned or not. Suppose, for instance, that there is $n=1$ feature function $f$, with $f(x)$ referring to the amount of tuning of $x$, and $\TT=\{x\in\X; \, f(x)\ge f_0\}$ the set of objects with a large amount of tuning (a special case of \eqref{fpfi} for $n=1$). Agent $\A$ wants to learn whether the proposition
$$
p: x_0\mbox{ is fine-tuned}
$$
is true or not. Data $\DD$ provides $\A$ with an estimate $\hmu=\hmu(\DD)$ of the amount of tuning of $x_0$. His posterior beliefs correspond to the Gibbs distribution (\ref{Gibbs2}), i.e.\ a density 
$
P(x) = P_0(x)e^{\lambda f(x)}/Z_\lambda
%\lb{PFT}
$
that is an exponentially tilted version of the prior density $P_0(x)$, with $\lambda=\lambda(\hmu)$. In \cite{DiazHossjer2022}, we considered algorithms whose outputs are drawn from this $P$. When $\lambda>0$, this algorithm generates outcomes in $\TT$, with a large amount of tuning, more often compared to chance, indicating that external knowledge has been infused into the algorithm. In our setting, $\TT$ is rather the truth set of proposition $p$. Moreover, $\P_0$ and $\P$ (with $\lambda>0$) correspond to beliefs of two agents $\I$ and $\A$, where $\A$ has stronger beliefs than $\I$ that the true structure $x_0$ is highly tuned. This framework has several applications. Firstly, if $\X$ is the set of values of a constant of nature, $f(x)$ quantifies the extent to which a value $x$ of this constant is consistent with a universe that harbors life. In \cite{DiazHossjerMathew2024}, we investigated whether it is possible to obtain LKA of a constant of nature being fine-tuned or not.  

Secondly, suppose $\X$ is a set of LLMs. Each LLM in $\X$ is first trained on broad data through self-supervision (a so called foundational model, cf. \cite{BommasaniEtAl2021}), but then adapted or fine-tuned on application-dependent data in order to more accurately perform specific tasks. In this context, $f(x)$ refers to the degree of adaptation or fine-tuning of LLM $x$. Agent $\A$ does not know $f(x_0)$, but he receives data from $x_0$ in order to test whether this data involves domain specific knowledge \cite{OrenEtAl2024}. This makes it possible for $\A$ to compute an estimate $\hmu$ of $f(x_0)$, and based on this he updates his prior beliefs about $x_0$ to $P(x)$, with $\lambda=\lambda(\hmu)$. An improved posterior could be derived by adding a second feature function $f^2(x)$ to take the variance of the estimate $\hmu$ into account (cf.\ Example \ref{Exa:Coord}).

\subsubsection{6.2.5 Using the true likelihood for primary learning from feature-based data} 

In our approach to LKA, $\A$'s posterior distribution minimizes the Kullback-Leibler divergence to $\I$'s prior, among all distributions that satisfy side constraints in terms of observed features. This can be viewed as a method of moments approach, where the observed moments of the features are used for inference of the posterior distribution. This approach implies that the likelihood \eqref{LikGibbs} of the posterior distribution is not the actual likelihood of data but rather a solution to an optimization problem. In contrast, in \cite{HossjerDiazRao2022} we used the true likelihood and defined the posterior distribution through Bayes Theorem. It would be interesting to combine ideas of the present article and \cite{HossjerDiazRao2022}, so that on one hand data $\DD$ are based on $n$ features, but on the other hand the true likelihood $L(\DD|x)$ of agent $\A$ is used in order to define his posterior distribution \eqref{Bayes2}.     

\subsubsection{6.2.6 Goodness of fit}

We have assumed that the true world $x_0$ belongs to the parameter set $\X$. A possible extension is to assume that $x_0\notin\X$. This happens, for instance, when $x_0$ is not among the set $\X$ of possible true world candidates of agent $\A$. Such an assumption would make it possible to define a goodness-of-fit test of whether the statistical model $\{L(\DD|x);\, x\in\X, \DD\in\Delta\}$ harbors $x_0$ or not. This is possible, not only within a frequentist framework, but also within a Bayesian framework \cite{Box1980, Rubin1984, GelmanShalizi2013}. But even when $x_0\notin\X$, there is typically one element $\hat{x}_0\in\X$ that is closest to $x_0$. With enough data points $N$, and sufficiently many features $n$, the posterior distribution of $\A$ will be close to a point mass at $\hat{x}_0$.  A related phenomenon occurs when $x_0\in\X$, but $\A$'s discernment is restricted to a $\s$-field that is generated from a countable partition $\PP=\{\AA_k; k=1,2,\ldots\}$ of $\X$. It may happen that $\A$ does not know the set $\X$. He is only aware of the elements of partition $\PP$ as atoms, but not the actual sets $\AA_k$ in $\X$ that these atoms correspond to. If $x_0\in\AA_{k_0}$ for some $k_0\ge 1$, $\AA_{k_0}$ takes the role of $\hat{x}_0$.

\begin{supplement}
The supplementary material \cite{DiazEtAlSupp2025} contains mathematical proofs of all the results in the main text.
\end{supplement}
%\begin{supplement}
%\stitle{Title of Supplement B}
%\sdescription{Short description of Supplement B.}
%\end{supplement}

\section*{Acknowledgement}

The authors wish to thank an associate editor and two anonymous reviewers, whose extensive comments considerably improved the quality of the paper. 
%VTeX will add %%  MR numbers.                      %%
%%                                                         %%
%%  Use \cite{...} to cite references in text.             %%
%%                                                         %%
%%%%%%%%%%%%%%%%%%%%%%%%%%%%%%%%%%%%%%%%%%%%%%%%%%%%%%%%%%%%%

%% if your bibliography is in bibtex format, uncomment commands:
\bibliographystyle{imsart-number} % Style BST file (imsart-number.bst or imsart-nameyear.bst)
\bibliography{/Users/daangapa/Documents/Research/daangapaBibliography.bib}       % Bibliography file (usually '*.bib')

%% or include bibliography directly:
%\begin{thebibliography}{4}
%%%
%\bibitem{r1}
%\textsc{Billingsley, P.} (1999). \textit{Convergence of
%Probability Measures}, 2nd ed.
%Wiley, New York.
%
%\bibitem{r2}
%\textsc{Bourbaki, N.}  (1966). \textit{General Topology}  \textbf{1}.
%Addison--Wesley, Reading, MA.
%
%\bibitem{r3}
%\textsc{Ethier, S. N.} and \textsc{Kurtz, T. G.} (1985).
%\textit{Markov Processes: Characterization and Convergence}.
%Wiley, New York.
%
%\bibitem{r4}
%\textsc{Prokhorov, Yu.} (1956).
%Convergence of random processes and limit theorems in probability
%theory. \textit{Theory  Probab.  Appl.}
%\textbf{1} 157--214.
%\end{thebibliography}

\end{document}

\begin{center}
    \LARGE
 {\bf Proofs of Results in ``Statistical Learning Does not Always Entail Knowledge''}
\end{center}

\section{Introduction}

In this supplementary material of the main article \cite{DiazEtAl2024}, we provide some additional examples and illustrations, as well as proofs of all results.

\section{Proofs of results from Section 2 of \texorpdfstring{\cite{DiazEtAl2024}}{Diaz et al. (2024)}}

{\bf On the formal construction of $\A$'s posterior beliefs}

Here we formalize the construction of the posterior beliefs of agent $\A$ in Section 2.1 of \cite{DiazEtAl2024}, based on data $\DD\in\Delta$. These data are used to update the beliefs of the ignorant person $\I$, a belief that corresponds to the distribution of the random variable $X\in\X$. To do so, we will assume that $\DD$ is an observation of a random variable $D$ taking values on some measurable space $(\Delta, \mathcal D)$. For some underlying sample space $\Omega$, we define the random element $(X, D): \Omega \to \X \times \Delta$ that is $(\F \times \mathcal D)$-measurable. Moreover, to the measurable product space $(\X \times \Delta, \F \times \mathcal D)$ we associate a joint law $\QQ^*$ with density $Q^*(x, \delta) = P_0(x) L(\delta \mid x)$ and marginal densities
\begin{align}\lb{Marginals}
	\int_{\X} Q^*(x, \delta)\D x = L(\delta), \quad\quad
	\int_{\Delta} Q^*(x, \delta)\D\delta = P_0(x).
\end{align}
Thus, the beliefs of $\I$ correspond to the density of $X$, whereas the posterior beliefs of agent $\A$ are obtained as the conditional density of $X$ given the event $\{D =\DD\}$, %With some slight abuse of notation, we call this density
expressed as
\begin{align}\lb{Post}
	P(x)  \defeq Q^*(x \mid \DD) 
	= \frac{Q^*(x, \DD)}{\int_\X Q^*(y, \DD)\D y}.
\end{align}
\qed

\begin{proposition}\lb{DiscFacts}
    	Let $\G_\A = \s(\AA_1, \AA_2, \ldots)$ be generated by a countable partition $\PP=\{\AA_1, \AA_2, \ldots\}$ of $\X$. If $\G_\I \subset \G_\A \subset \G 	\subset \F$, the following follows:
 	\begin{enumerate}[label = (\arabic*)]
        		\item\lb{Fact1} If $\AA \in \G_\I$, then $\P_0\infdivx{\AA}{\G_\I} = \P\infdivx{\AA}{\G_\A} = \1_\AA$, a.s.
        		\item\lb{Fact2} If $\AA \in \G_\A \setminus \G_\I$,  then $\1_\AA = \P\infdivx{\AA}{\G_\A} =  \P_0\infdivx{\AA}{\G}$, a.s.
        		\item\lb{Fact3} The function $\E_\P\infdivx{g}{\G_\A}$ is piecewise constant over all sets $\AA_i \in \PP$ that 			generate $\G_\A$. If additionally $\P(\AA_i) \neq \P_0(\AA_i)$ and $\E_\P\infdivx{g}{\G_\A}$ is nonzero on $\AA_i$, 			then $\int_{\AA_i} \E_\P\infdivx{g}{\G_\A} \D\P \neq \int_{\AA_i} \E_{\P_0}\infdivx{g}{\G_\A} \D \P_0$.
        		\item\lb{Fact4} $\P(\TT) = \E_\P[\E_{\P_0}\infdivx{f_p}{\G_\A}]$.
        		\item\lb{Fact5} If $\G_\I = \{ \emptyset, \X\}$, $\P_0(\TT) = \E_{\P_0} \infdivx{f_p}{\G_\I}$ a.s.
    	\end{enumerate}
\end{proposition}

Before proving Facts \ref{Fact1}-\ref{Fact5} of Proposition \ref{DiscFacts}, let us first comment on them. The first part of Fact \ref{Fact1} ($\P_0\infdivx{\AA}{\G_\I} = \1_\AA$) implies that the ignorant agent $\I$, within his lower discernment $\G_\I$, has the {\sl potential} of knowing with certainty whether an event $\AA\in\G_I$ happened (i.e.\ $x_0\in\AA$) or not, by appropriate choice of $\P_0$. Consequently, Fact \ref{Fact1} implies that if $\I$ has the potential to know $\AA$ with certainty, so does agent $\A$ with his additional discernment. Fact \ref{Fact2} says that had the ignorant agent $\I$ at least the same discernment as $\A$, he would have the {\sl potential} to know with certainty whether any event $\AA\in\G_\A$ within $\A$'s discernment happened or not. Fact \ref{Fact3} says that, despite the LHS and RHS of equation (6) of \cite{DiazEtAl2024} being equal with probability 1, their integrals with respect to $\P$ and $\P_0$ can be different. Together with Fact \ref{Fact2}, it says that the conditional probability function of $\AA$ can have different integrals under $\P$ than under $\P_0$. Facts \ref{Fact4} and \ref{Fact5} are applications of the tower property. 

\begin{proof}
For $\AA \in \F$, let $g \defeq \1_\AA$. Then Definition 2.1 of \cite{DiazEtAl2024} implies that
\beq
    \P\infdivx{\AA}{\G} = \P_0\infdivx{\AA}{\G},
\lb{CntEx0}
\eeq
a.s. To prove Fact \ref{Fact1}, assume $\AA \in \G_\I$. Then
\beq
    \1_\AA = \P_0\infdivx{\AA}{\G_\I} = \P_0\infdivx{\AA}{\G} = \P\infdivx{\AA}{\G} = \P\infdivx{\AA}{\G_\A},  
\lb{CntEx1}
\eeq
a.s., where the first equality is due to the fact that $\1_\AA$ is a version of $\P_0\infdivx{\AA}{\G_\I}$; the second equality is due to the fact that $\AA \in \G_\I \Rightarrow \AA \in \G$;  the third equality is due to \eqref{CntEx0}; and  the last equality is due to the fact that $\AA \in \G_\A \subset \G$, since $\AA \in \G_\I$. Moreover, the first and third equalities in \eqref{CntEx1} are a.s.

To prove Fact \ref{Fact2}, assume $\AA \in \G_\A \setminus \G_\I$. Then \eqref{CntEx0} implies that
\beq
    \1_\AA = \P\infdivx{\AA}{\G_\A} = \P\infdivx{\AA}{\G} = \P_0\infdivx{\AA}{\G}.
\eeq

To prove Fact \ref{Fact3}, let $c_i$ be the constant value of $\E_\P\infdivx{g}{\G_\A}= \E_{\P_0}\infdivx{g}{\G_\A}$ on $\AA_i$. Then, since $\P(\AA_i)\ne \P_0(\AA_i)$, if $c_i\ne 0$ it follows that
\begin{align}\label{Discernment2}
	\int_{\AA_i} \E_\P\infdivx{g}{\G} \D\P = c_i \P(\AA_i) \ne c_i \P_0(\AA_i) = \int_{\AA_i} \E_{\P_0}\infdivx{g}{\G} \D \P_0.
\end{align}

As for Fact \ref{Fact4}, it was proven in \cite{HossjerDiazRao2022}, but we present its proof here for completion:
\begin{align}\lb{CntEx2}
    \P(\TT) = \E_\P(f_p) = \E_\P[\E_\P\infdivx{f_p}{\G_\A}] = \E_\P[\E_{\P_0}\infdivx{f_p}{\G_\A}],        
\end{align}
where the first equality is obtained by definition of $f_p$, the second is an application of the tower property, and the last one uses the discernment property (6) of \cite{DiazEtAl2024}.

To prove Fact \ref{Fact5} observe that if $\G_\I=\{\emptyset, \X\}$, then $\E_{\P_0}\infdivx{f_p}{\G_\I}$ is constant a.s. The result then follows from a second application 
$$
\P_0(\TT) = \E_{\P_0}(f_p) = \E_{\P_0}[\E_{\P_0}\infdivx{f_p}{\G_\I}] = \E_{\P_0}\infdivx{f_p}{\G_\I}
$$
of the tower property, with the last identity holding a.s. 
\end{proof}

\begin{example}[Countable and cocountable sets.]\lb{Cocon}
This example shows that the discernment $\s$-field $\G_\A$ of agent $\A$, according to Definition 2.1 of \cite{DiazEtAl2024}, cannot always be extended to $\s$-fields that are not generated from a countable partition. Our example is based on the following example presented by Billingsley \cite[Example~33.11]{Billingsley1995}: Consider the probability space $(\X,\F,\QQ)$, where $\X=[0,1]$, $\F$ is the Borel $\s$-field on $[0,1]$, and $\QQ$ a continuous probability measure. Consider an agent $\A$ whose discernment $\G_\A$ is given by the countable-cocountable subsets of $[0,1]$ (i.e., $\BB \in \G_\A$ if and only if either $\BB$ is countable or $\BB^c$ is countable). Then, for all $\AA \in \F$, 
\beq
	\QQ(\AA)  = \QQ\infdivx{\AA}{\G_\A}, 
	\lb{Cocon1}
\eeq
a.s. Indeed, since $\QQ\infdivx{\AA}{\G_\A}$ is $\G_\A$-measurable and integrable, it follows that
\beq
	\int_\BB \QQ(\AA) \QQ(\D x) = \QQ(\AA) \QQ(\BB) = \QQ(\AA \cap \BB)
    = \int_\BB \QQ\infdivx{\AA}{\G_\A}(x) \QQ(\D x)    
\lb{Qint}
\eeq
for all $\BB \in \G_\A$. This is so since both sides of \eqref{Qint} are either 0 or $\QQ(\AA)$, depending on whether $\BB$ or $\BB^c$ is countable. And by the definition of conditional expectation, \eqref{Cocon1} follows from \eqref{Qint}. 
On the other hand, since every singleton of $[0,1]$ is $\G_\A$-measurable, seeing $\G_\A$ as discernment, we would intuitively expect that
\beq
	\1_\AA = \QQ\infdivx{\AA}{\G_\A},
	\lb{Cocon2}
\eeq
since $\AA$ is a union of singletons. However, this intuition goes wrong whenever $\QQ(\AA)>0$, so that the union is uncountable. Indeed, taking \eqref{Cocon2} together with \eqref{Cocon1}, we obtain $\QQ(\AA) = \QQ\infdivx{\AA}{\G_\A} = \1_\AA$, a contradiction for all $\AA$ such that $0<\QQ(\AA)<1$. 

Suppose Definition 2.1 in \cite{DiazEtAl2024} holds and  $\P_0$ is the uniform distribution on $\X=[0,1]$. Then by Bayes Theorem $\P$ must also have a continuous distribution on $\X$. In addition, for any $\AA\in\F$, it follows from (6) of \cite{DiazEtAl2024} (with $g=\1_{\AA}$) and \eqref{Cocon1}, applied to $\P_0$ and $\P$, that
$$
\P_0(\AA) = \P_0\infdivx{\AA}{\G_\A}
= \P\infdivx{\AA}{\G_\A} = \P(\AA).
$$
Since $\AA\in\F$ is arbitrary, we conclude that $\P_0=\P$. Consequently, when (6) of \cite{DiazEtAl2024} and a maxent uniform prior are assumed, we obtain the unreasonable result that the posterior cannot differ from the prior.
\qed
\end{example}

\begin{theorem}\lb{FLFK}
    For the topological space $(\X, \mathcal O)$, consider the measurable space $(\X, \F)$, where $\F = \s(\mathcal O)$. Let $\P_0$ be a %fully supported 
    probability measure on $(\X, \F)$ and define another probability measure $\P$ on $(\X, \F)$ as in eq.~(4) of \cite{DiazEtAl2024}, where $\P_0$ and $\P$ represent beliefs about the true world $x_0\in\X$ of two agents $\I$ and $\A$ respectively. Assume that $\P_0$ and $\P$ are measurable with respect to $\s$-fields $\G_\I$ and $\G_\A$ on 	$\X$, with $\G_\I \subsetneq \G_\A \subset \F$. Assume further that $\G_\A = \s(\PP)$ is generated from a countable partition $\PP = \s(\AA_1, \AA_2, \ldots)$ such that $\P_0(\AA_i) > 0$ for all $\AA_i \in \PP$ and none of the $\AA_i \in \PP$ is $\G_\I$-measurable. Let $p$ be a proposition that is true in a set of worlds $\TT \in \F$. Then
	\begin{enumerate}[nosep, label = \roman*.]
		\item\lb{FuLeT} If for all $\AA \in \PP$, it holds that $\AA \not \subset \TT$ and $\P_0(\AA \setminus \TT) >0$, then $\P(\TT) < 1$. In particular, if $p$ is true in the true world $x_0$, this implies that full learning of $p$ is not possible.  
		\item\lb{FuLeTT} Suppose \ref{FuLeT} fails in the sense that there is an $\AA \in \PP$ such that $\AA \subset \TT$. Then we can choose $x_0$ so that $p$ is true in $x_0$, and $\P$ according to eq.~(5) in \cite{DiazEtAl2024}, so that there is full learning of $p$, i.e.\ $\P(\TT) = 1$.
		\item\lb{FuLeF} If for all $\AA \in \PP$, it holds that $\TT\cap\AA\ne\emptyset$ and $\P_0(\TT\cap\AA) >0$, then $\P(\TT) >0$. In particular, if $p$ is false in the true world $x_0$, this implies that full learning of $p$ is not possible. 
		\item\lb{FuLeFF} Suppose \ref{FuLeF} fails in the sense that there is $\AA \in \PP$ such that $\AA \cap \TT = \emptyset$. Then we can choose $x_0$ such that $p$ is false in $x_0$, and $\P$ according to eq.~(5) in \cite{DiazEtAl2024}, so that there is full learning of $p$, i.e.\ $\P(\TT) = 0$.
		\item\lb{FuKn} If there is $\AA \in \PP$ such that $\{x_0\} \subsetneq \AA$ and $\P_0(\AA \setminus \{x_0\}) > 0$, then 			$\P(\{x_0\}) < 1$ and full knowledge acquisition of not possible.
		\item\lb{FuKnn} If $\{x_0\} \in \PP$, then it is possible to choose $\P$ according to eq.~(5) in \cite{DiazEtAl2024} such that $\P(x_0) = 1$.
	\end{enumerate}
\end{theorem}

\begin{proof} All six parts i-vi of the theorem are proven in order: 
\begin{enumerate}[nosep, label = \roman*.]
\item 
For each set $\AA_i$ of the partition $\PP$, define
\beq
q_i = \P(\TT|\AA_i) = \P_0(\TT|\AA_i) = 1 - \frac{\P_0(\AA_i\setminus \TT)}{\P_0(\AA_i)} < 1,
\lb{qi}
\eeq
where the last step is a consequence of the assumptions $\P_0(\AA_i)>0$ and $\P_0(\AA_i\setminus \TT)>0$. It follows from the Law of Total Probability that 
$$
\P(\TT) = \sum_i \P(\AA_i)q_i < \sum_i \P(\AA_i) = 1,
$$
where the inequality was deduced from \eqref{qi} and the fact that $\P(\AA_i)>0$ for at least one $i$. 
%Without loss of generality, assume that $\TT \subsetneq \AA$ for some $\AA \in \PP$. Then
%	\begin{align}\lb{ACT}
%	\begin{aligned}
%		\P(\TT) &= \P(\TT \cap \AA) \\
%				&= \int_{\AA} \P \infdivx{\TT}{\G_\A} \D \P \\
%				&= \int_{\AA} \E_\P \infdivx{\1_\TT}{\G_\A} \D \P \\
%				&= \int_\AA \E_{\P_0} \infdivx{\1_\TT}{\G_\A} \D \P \\
%				&= \int_\AA \P_0 \infdivx{\TT}{\G_\A} \D \P \\
%				&= \P_0 (\TT \mid \AA)\P(\AA) \\
%                &\le \P_0(\TT\mid\AA)\\
%                &= 1 - \frac{\P_0(\AA\setminus \TT)}{\P_0(\AA)}\\
%				&< 1,
%	\end{aligned}
%	\end{align}
%	where the fourth equality uses \eqref{Discernment}, whereas in the last inequality we employed the assumption 			$\P_0(\AA\setminus \TT) >0$ of Part \ref{FuLeT}. 
	
	\item 
	If $i_0$ is the index for which $\AA_{i_0} \subset \TT$, choose $x_0\in \AA_{i_0}$ and $\P(\AA_{i_0})=1$.
	
	\item
	Note that $\TT^c = \X\setminus\TT$ satisfies the conditions of Theorem \ref{FLFK}.\ref{FuLeF}. Hence $\P(\TT^c)<1$ and 		$\P(\TT)=1-\P(\TT^c) > 0$. 
	
	\item
	If $i_0$ is the index for which $\P_0(\AA_{i_0}\cap \TT)=0$, choose $x_0\in \AA_{i_0}$ and $\P(\AA_{i_0})=1$.
	
	\item
	Make $\TT = \{x_0\}$ in Theorem \ref{FLFK}.\ref{FuLeT}. The result follows.
	
	\item
	This is trivial.
	\end{enumerate}
\end{proof}

\section{Proofs of results from Section 3 of \texorpdfstring{\cite{DiazEtAl2024}}{Diaz et al. (2024)}}

{\bf Motivation that the Gibbs distribution solves the constrained minimization problem (10)-(11) of \cite{DiazEtAl2024}}

In order to motivate that the Gibbs distribution density
\begin{align}\label{Gibbs2}
	P(x) = Q_{\boldsymbol{\lambda}}(x) = \frac{P_0(x) e^{\boldsymbol{\lambda} \cdot \mathbf f(x)}}{Z_{\boldsymbol\lambda}}
\end{align}
is the solution to the minimization problem (10)-(11) of \cite{DiazEtAl2024}, we will use Lagrange multipliers. Our goal is to find the distribution $\QQ\in\QQQ$ that minimizes the loss function
\beq
\L(\QQ) = \int_\X Q(x) \left[ \log \frac{Q(x)}{P_0(x)} - \bla \cdot \f(x) - \xi \right] \D x 
- (\bla \cdot \bmu - \xi), 
\lb{LQ}
\eeq
where $\bmu = (\mu_1,\ldots,\mu_n)^T$ are the features that the expected features $\bmu(\QQ)$ of $\QQ$ must equal. The minimizer of (\ref{LQ}) must satisfy 
$$
0 = \frac{\partial \L(\QQ)}{\partial Q(x)} 
= \log \frac{Q(x)}{P_0(x)} + 1 - \bla\cdot \f(x) - \xi
$$
for all $x\in\X$, with solution 
\beq
Q(x) = P_0(x) \exp(\bla \cdot \f(x) + \xi -1).
\lb{QQx}
\eeq
The constants $\bla$ and $\xi$ are chosen in (\ref{QQx}) so that the side constraints $\mu_i(\QQ)=\mu_i$, $i=1, \ldots, n$, and $\int_\X Q(x)\D x = 1$ are fulfilled, and this is equivalent to (\ref{Gibbs2}). \hfill $\Box$

\begin{proposition}\lb{Prop:PAsympt}
Let $\P=\P_{\hbmu(\DD)}$ refer to the solution of the optimization problem 
\beq
\P = \P_{\hbmu(\DD)} = \mbox{arg} \inf_{\QQ\in\QQQ(\hbmu)} \infdiv{\QQ}{\P_0}, 	
\lb{Popt}
\eeq
with an estimated feature vector $\hbmu=\hbmu(\DD)$ that is an observation of the random vector $\hbmu(D)$. Assume that data $D=(D_1,\ldots,D_N)$ consists of $N$ data items, and that convergence in probability
\beq
\hbmu(D)\stackrel{p}{\to} \f(x_0)
\lb{muConvLargeN}
\eeq
holds as $N\to\infty$, where $x_0$ is the true but unknown value of $x$. Then  
\beq
\P_{\hbmu(D)} \stackrel{\mathcal L}{\to} \P_\infty
\lb{PConvLargeN}
\eeq
as $N\to\infty$ with probability 1, where $\P_\infty$ is the Gibbs distribution \eqref{Gibbs2} with $\bmu(\P_\infty)=\f (x_0)$.
\end{proposition}

\begin{proof}
Write $P_{\bmu}(x)=P(x;\bmu)$ for the probability function or density function of the solution $\P$ to the optimization problem \eqref{Popt}, and let $\bmu_\infty = \f(x_0)$ for the limiting value of $\hbmu=\hbmu(\DD)$ in \eqref{muConvLargeN} as $N\to\infty$. When $\P$ is a discrete distribution we have that 
\beq
\P(\AA;\bmu) = \sum_{x\in\AA} P(x;\bmu) 
\lb{PAmu}
\eeq
for each $\AA\in\F$. Since $0\le P(x;\bmu)\le 1$ and $\bmu\to P(x;\bmu)$ is a continuous function for each $x\in\X$ it follows from the Dominated Convergence Theorem that $\P(\AA;\bmu)\to \P(\AA;\bmu_\infty)=\P_\infty(\AA)$ as $\bmu\to\bmu_\infty$ for each $\AA\in\F$. Invoking \eqref{muConvLargeN} we find that $\P(\AA;\hbmu(D))\stackrel{p}{\to} \P_\infty(\AA)$ for each $\AA\in\F$. In particular, we have 
$\P(\{x\};\hbmu(D))\to \P_\infty(\{x\})$ as $N\to\infty$ with probability 1 for all $x\in\X$, 
proving \eqref{PConvLargeN}.  
\end{proof}

\begin{theorem}\lb{Thrm:PAsympt}
Assume that the estimates features $\hbmu(\DD)$ are obtained from an independent sample $\DD=(\DD_1,\ldots,\DD_N)$ as a sample average
\beq
\hbmu = \hbmu(\DD) = \frac{1}{N} \sum_{k=1}^N \hbmu(\DD_k)
\lb{muDDInd}
\eeq
where $\{\hbmu(\DD_k)\}_{k=1}^N$ are observations of independent and identically distributed random variables $\hbmu(D_k)$. Assume also that the estimated features $\hbmu(D_k)$ for all data items $D_k$ are unbiased with a finite second moment, i.e.\ $E[\hbmu(D_k)]=\f(x_0)$ and $\Var[(\hbmu(D_k)]=\bSi$, where $\bSi$ is a covariance matrix of order $n$. 
We then have weak convergence
\beq
\sqrt{N}(\hbmu(D)-\f(x_0)) \stackrel{\mathcal L}{\to} N(0,\bSi)
\lb{muDDConv}
\eeq
as $N\to\infty$. In addition
\beq
\sqrt{N}(\P_{\hbmu(D)}-\P_\infty) \stackrel{\mathcal L}{\to} \W
\lb{PDDConv}
\eeq
as $N\to\infty$ with probability 1, where $\P_{\hbmu(D)}$ is defined as in Proposition \ref{Prop:PAsympt}, $\P_\infty$ is defined below \eqref{PConvLargeN}, whereas $\W$ is a Gaussian signed measure on $\X$, with $\W(\AA)\sim N(0,C(\AA,\AA))$ and $\Cov(\W(\AA),\W(\BB))=C(\AA,\BB)$ for all $\AA,\BB\in\F$, and with $C(\AA,\BB)$ is defined in the proof below.  
\end{theorem}

\begin{proof}
Equation \eqref{muDDConv} follows directly from the Central Limit Theorem. In order to prove \eqref{PDDConv}, we follow that proof of Proposition \ref{Prop:PAsympt} and write $P_{\bmu}(x)=P(x;\bmu)$ for the probability function or density function of the solution $\P$ to the optimization problem \eqref{Popt}. Let also $\bmu_\infty = \f(x_0)$ be the limiting value of $\hbmu=\hbmu(\DD)$ in \eqref{muDDConv} as $N\to\infty$. Suppose that $\P$ is a discrete distribution. For each $\AA\in\F$ we then use the Delta method, that is, a first-order Taylor expansion of \eqref{PAmu} around the point $\bmu_\infty$, according to
$$
\P(\AA;\bmu) \approx \P(\AA;\bmu_\infty) + \P^\prime(\AA;\bmu_\infty) (\bmu-\bmu_\infty)^T. 
$$
Here $\P(\AA;\bmu_\infty)=\P_\infty(\AA)$, $\P^\prime(\AA;\bmu)= \D \P(\AA;\bmu)/\D \bmu$, whereas $T$ refers to vector transposition. Then \eqref{PDDConv} follows from \eqref{muDDConv}, with 
$$
C(\AA,\BB) = \P^\prime(\AA;\bmu_\infty) \bSi\P^\prime(\BB;\bmu_\infty)^T.
$$
\end{proof}

\section{Proofs of results from Section 4 of \texorpdfstring{\cite{DiazEtAl2024}}{Diaz et al. (2024)}}

\begin{proposition}\lb{Prop:Learnfi}
	Consider a proposition $p$ which is true in the set of worlds 
    \beq
	\TT = \{x \in \X : f_i(x)\ge f_0 \}.
	%\lb{fpfi}
\eeq
    Assume further that
	\beq
		\min_{x\in\X} f_i(x) \le f_0 \le \max_{x\in\X} f_i(x),
	\lb{Ineq}
        \eeq
	with at least one of the two inequalities being strict. Then $\P(\TT) = \QQ_{\bla}(\TT)$ is a strictly increasing function of $\lambda_i$, with 
    \beq
    \begin{array}{rcl}
    \lim_{\lambda_i\to -\infty} \QQ_{\bla}(\TT) &=& 0,\\
     \lim_{\lambda_i\to \infty} \QQ_{\bla}(\TT) &=& 1    \end{array}
    \lb{PlambdaLimit}
    \eeq
    when the other $n-1$ components of $\bla$ are kept fixed. 
    In particular, agent $\A$ learns $p$ (in relation to the ignorant person $\I$), if the two conditions below hold:
	\begin{enumerate}[label=(\roman*)]
		\item $\lambda_j=0$ for all $j \in \{1,\ldots,n\} 	\setminus \{i\}$, 
		\item either $\lambda_i>0$ and $f(x_0)\ge f_0$, or $\lambda_i<0$ and $f(x_0)<f_0$.  
	\end{enumerate}     
\end{proposition}

\begin{proof}
In order to verify that $\P(\TT)=\QQ_{\bla}(\TT)$ is a strictly increasing function of $\lambda_i$, we use the same method of proof as in Proposition 1 of \cite{DiazHossjer2022}. To this end it is convenient to introduce $\tilde{\P}=\QQ_{\tilde{\bla}}$, where $\tilde{\bla}=(\tilde{\lambda}_1,\ldots,\tilde{\lambda}_n)$ has components
$$
\tilde{\lambda}_j = \left\{\begin{array}{ll}
\lambda_j; & j\ne i,\\
0; & j=i.
\end{array}\right.
$$
Let also $\tilde{P}(x)$ be the probability function or density of $\tilde{\P}$, when $\X$ is countable and continuous respectively. Define
	\begin{equation}
		\begin{aligned}\label{JK}
			J(\lambda_i) &= \sum_{x\in \TT^c} e^{\lambda_i[f(x)-f(x_0)]}\tilde{P}(x),\\
			K(\lambda_i) &= \sum_{x\in \TT} e^{\lambda_i[f(x)-f(x_0)]}\tilde{P}(x),
		\end{aligned}
	\end{equation}
	when $\X$ is countable, and replace the sums in \eqref{JK} by integrals when $\X$ is continuous. Then 
	\begin{align}\label{PJK}
		\QQ_{\bla}(\TT) &=  \frac{e^{\lambda_i f(x_0)}K(\lambda_i)}{ e^{\lambda_i f(x_0)}[J(\lambda_i)+K(\lambda_i)]} \nonumber \\
		&= \frac{K(\lambda_i)}{J(\lambda_i) + K(\lambda_i)} \\
		&= \frac{1}{\frac{J(\lambda_i)}{K(\lambda_i)} +1}. \nonumber
	\end{align}

	Since by assumption $f_0$ is an interior point of the range of $f_i$, it follows that $0< \tilde{\P}(\TT)<1$. From this, we deduce that $J(\lambda_i)$ is a strictly decreasing function of $\lambda_i$, and/or $K(\lambda_i)$ is a strictly increasing function of $\lambda_i$. This implies that $\P(\TT)=\QQ_{\bla}(\TT)$ is a strictly increasing function of $\lambda_i$. The lower part of \eqref{PlambdaLimit} follows from the fact that
    \beq
    \begin{array}{rcl}
    \lim_{\lambda_i\to\infty} J(\lambda_i) &=& 0,\\
    \lim_{\lambda_i\to\infty} K(\lambda_i) &=& \infty
    \end{array}
    \lb{JKLim}
    \eeq
    when both inequalities of \eqref{Ineq} are strict. If only one of the two inequalities of \eqref{Ineq} is strict, then at least one of the two limits of \eqref{JKLim} are valid, so that \eqref{PlambdaLimit} still holds. The upper part of \eqref{PlambdaLimit} is proved similarly. 
    
    The second part of Proposition \ref{Prop:Learnfi} then follows from the definition of learning in Definition 2.2 of \cite{DiazEtAl2024}, and the facts that $\P=\QQ_{\bla}$ and $\tilde{\P}=\QQ_{\tilde{\bla}}=\P_0$ when $\tilde{\bla}=(0,\ldots,0)$. 
\end{proof}

\begin{theorem}[Fundamental limits of knowledge]\lb{Prop:FinPop}
Consider a finite set $\X=\{x_1,\ldots,x_d\}$ with $n$ binary features 
\beq
	f_i(x) = \1_{\AA_i}(x)
	\lb{fiFinite}
\eeq
that are indicator functions for different subsets $\AA_1,\ldots,\AA_n$ of $\X$. %Assume that $\A$'s posterior beliefs \eqref{PFinite} are based on the $n$ binary features \eqref{fiFinite}. 
If 
\beq
	n\ge \lceil \log_2 d \rceil,
	\lb{nlow}
\eeq
it is possible to choose the sets $\AA_1,\ldots,\AA_n$ and constants $\lambda_1,\ldots,\lambda_n$ so that full knowledge can be attained about any proposition $p$. Conversely, if $n$ does not satisfy \eqref{nlow}, for any choice of $n$ binary features, it is possible to pick $x_0$ so that full knowledge acquisition is not possible. 
\end{theorem}

\begin{proof}  The Gibbs distribution $\P$ in \eqref{Gibbs2} has a probability function 
\beq
	P(x_k) = \frac{\exp\left[\sum_{i=1}^n \lambda_i \1_{\AA_i}(x_k)\right]}{\sum_{l=1}^d \exp\left[\sum_{i=1}^n \lambda_i 				\1_{\AA_i}(x_l)\right]}
	\lb{PFinite}
\eeq
for some constants $\lambda_1,\ldots,\lambda_n$ that quantify the impact of each feature on agent $\A$'s posterior beliefs.
In this case, data $\DD \in \Delta$ provide $\A$ with information about the probability $\hmu_i = \E_\P f_i(X) = \P(\AA_i)$ of each set $\AA_i$, so that $\P = \P_{\hbmu}$.   

We will first show that whenever (\ref{nlow}) holds, there are feature functions $f_1,\ldots,f_n$ in (\ref{fiFinite}) such that for any $x\in\X$ it is possible to choose the parameter vector $\bla = \bla_x$ of the Gibbs distribution $\P$ in (\ref{PFinite}), that represents agent $\A$'s beliefs, so that $\P=\bde_x$ is a point mass at $x$ and hence $P(x) = 1$. This will prove the result since, in particular for the true world $x_0$, it implies that
\beq
	P(x_0) = 1.
	\lb{FullK}
\eeq
is equivalent to full knowledge acquisition of $\A$ for any proposition $p$ (see Definition 2.3 of \cite{DiazEtAl2024}). With $n$ as in (\ref{nlow}) it is possible to write $x_k=(x_{k1},\ldots,x_{kn})\in \X$ as a binary expansion of the number $k-1$ for $k=1,\ldots,d$. Then choose the indicator sets of the feature functions (\ref{fiFinite}) as
$$
\AA_i = \{x_k; x_{ki} = 1\}
$$
for $i=1,\ldots,n$. Let $x_0=(x_{01},\ldots,x_{0n})$ be the binary expansion of $x_0=x_{k_0}$, and let $\lambda > 0$ be a large number. Pick $\bla=\bla_{x_0}=(\lambda_1,\ldots,\lambda_n)$ so that
$$
\lambda_i = \left\{\begin{array}{ll}
\lambda; & \mbox{if }x_{0i}=1,\\
-\lambda; & \mbox{if }x_{0i}=0.
\end{array}\right. 
$$
For each $x_k\in\X$ we define the two subsets $I_0(x_k)=\{i; x_{ki}=0\}$ and $I_1(x_k)=\{i;x_{ki}=1\}$ of $\{1,\ldots,n\}$. It follows from (\ref{PFinite}) that 
$$
P(x_k) = C e^{\lambda n_k} 
$$
where $n_k=|I_1(x_0)\cap I_1(x_k)|-|I_0(x_0)\setminus I_0(x_k)|$ is an integer and $C$ is a normalizing constant assuring that $\P$ is a probability measure. Since $k\in\{1,\ldots,d\}\to n_k$ is uniquely maximized for $k=k_0$ by $n_{k_0}=|I_1(x_0)|$, equation \eqref{FullK} follows by letting $\lambda\to\infty$. This completes the proof of the first part of Proposition \ref{Prop:FinPop}. 

Assume next that \eqref{nlow} does not hold, so that $n<\log_2 d$ and $2^n< d$. For each binary vector $\f=(f_1,\ldots,f_n)$ of length $n$, define the set
\beq
    \BB_{\f}=\{x\in\X; \, \f(x)=(f_1(x),\ldots,f_n(x))=\f\}.
    \lb{Bi}
\eeq
Suppose $d_0 \le 2^n$ of the $2^n$ sets in \eqref{Bi} are non-empty. It follows from \eqref{PFinite} that agent $\A$'s posterior probability function $P(x)$ is constant on each non-empty set in \eqref{Bi}. Since these $d_0$ non-empty sets form a disjoint decomposition of $\X$, and $d_0\le 2^n < d$, it follows that $|\BB_{\f_0}|>1$ for at least one binary vector $\f_0$. If $x_0\in \BB_{\f_0}$ we deduce that $P(x_0;\bla)\le 1/|\BB_{\f_0}| \le 0.5$, regardless of the value of $\bla$. According to Definition 2.3 of \cite{DiazEtAl2024}, full knowledge acquisition is not possible for this particular $x_0$.      
\end{proof}

\begin{theorem}\lb{Prop:Coord} 
	In the setting of Example 3 of \cite{DiazEtAl2024}, consider propositions $p$ with
	\beq
		\TT = \{x\in [0,1]^n; f_p(x)=1\} = \times_{i=1}^n [a_i,b_i],
		\lb{TTCoord}
	\eeq
	where $0\le a_i < b_i \le 1$ for $i=1,\ldots,n$. For propositions $p$ that satisfy \eqref{TTCoord} and are true ($x_0\in\TT$), it is possible for $\A$ to come arbitrarily close to full learning if and only if at least one of the two conditions $a_i=0$ and $b_i=1$ holds for each 			$i=1,\ldots,n$. Moreover, it is only possible for $\A$ to come arbitrarily close to full knowledge about $p$ if, additionally, {\it all} coordinates of $x_0$ are either 0 or 1.   
\end{theorem}

\begin{proof}
Recall that $\A$ forms his beliefs according the Gibbs distribution with density
\beq
	P(x)=\prod_{i=1}^n P_i(x_i),
\lb{PCoord}
\eeq
where
\beq
	P_i(x_i) = 
	\left\{\begin{array}{ll}
		1, & \lambda_i = 0,\\
		\frac{\lambda_i e^{\lambda_i x_i}}{e^{\lambda_i}-1}, & \lambda_i\ne 0.
	\end{array}\right. 
\lb{PiCoord}
\eeq
for some vector $\bla=(\lambda_1,\ldots,\lambda_n)$, and that the set $\TT$ of worlds for which the proposition $p$ is true is given by (\ref{TTCoord}). Since we assume that $p$ is true ($x_0\in\TT$), it follows from Definition 2.2 of \cite{DiazEtAl2024} that it is possible to come artibrarily close to full learning of $p$ if for any $\epsilon>0$ we can find a vector $\bla=\bla_\epsilon$ such that 
\beq
	\P(\TT;\bla) \ge 1 - \epsilon.
	\lb{PTTla}
\eeq
Thus, we need to look more closely at $\P(\TT;\bla)$. Equations \eqref{TTCoord}--\eqref{PiCoord} imply that
\beq
	\P(\TT;\bla) = \prod_{i=1}^n \int_{a_i}^{b_i} P_i(x_i)\D x_i = \prod_{i=1}^n G(a_i,b_i;\lambda_i),
	\lb{PTTla2}
\eeq
where 
$$
G(a,b,\lambda) = \left\{\begin{array}{ll}
	\frac{e^{\lambda b}- e^{\lambda a}}{e^\lambda-1}; & \mbox{if }\lambda\ne 0,\\
b-a; & \mbox{if }\lambda=0. 
\end{array}\right.
$$
Maximizing (\ref{PTTla2}) with respect to $\bla$, it can be seen that
\beq
	\sup_{\bla} \P(\TT;\bla) = \prod_{i=1}^n \bar{G}(a_i,b_i),
	\lb{supPTTla}
\eeq
where
\beq
\bar{G}(a,b) = \sup_{\lambda} G(a,b;\lambda)  
\begin{cases}
=1; & \mbox{if at least one of $a=0$ or $b=1$ holds},\\
< 1; & \mbox{otherwise}.
\end{cases}
\lb{barG}
\eeq
We deduce from (\ref{supPTTla})-(\ref{barG}) that 
\beq
	\sup_{\bla} \P(\TT;\bla) = 1
	\lb{supPTTla2}
\eeq
if and only if at least one of the two conditions $a_i=0$ or $b_i=1$ holds for $i=1,\ldots,n$. In view of (\ref{PTTla}), this proves the first (learning) part of the theorem.  

We also need to verify the stated conditions on the true world $x_0=(x_{01},\ldots,x_{0n})\in \TT$ that make it possible for $\A$ to come arbitrarily close to full knowledge acquisition about $p$. In view of Definition 2.3 of \cite{DiazEtAl2024}, we must verify that  
\beq
\sup_{\bla} \P(B_\epsilon(x_0);\bla) = 1
\lb{PepsKA}
\eeq
for any ball $B_\epsilon(x_0)$ of radius $\epsilon>0$ surrounding $x_0$. Since each marginal density $P_i$ in (\ref{PiCoord}) is monotone in $x_i$, it is clear that (\ref{PepsKA}) holds only if for each $i=1,\ldots,n$, either $x_{0i}=0$ or $x_{0i}=1$, with the maximum in (\ref{PepsKA}) being attained in the limit where $\lambda_i\to -\infty$ if $x_{0i}=0$ and $\lambda_i\to\infty$ if $x_{0i}=1$ respectively.  
\end{proof}

\begin{theorem}\lb{Thrm:Coord}
In the setting of Example 4 of \cite{DiazEtAl2024}, it is possible, by appropriate choice of $\bla$, to come arbitrarily close to full learning and full knowledge of any proposition $p$ such that either a) $p$ is true and $x_0$ is an interior point of the truth set $\TT$, or b) $p$ is false and $x_0$ is an interior point of $\TT^c$.       
\end{theorem}

\begin{proof}
Assume without loss of generality that $p$ is true (the proof is analogous when $p$ is false) and that the supremum norm $d(x,y)=\max_{1\le i \le n/2} |x_i-y_i|$ is used as a distance between the elements of $\X$. Since, by assumption, $x_0=(x_{01},\ldots,x_{0,n/2})$ is an interior point of $\TT$, we can choose $\varepsilon >0$ so small that the closed ball of radius $\varepsilon$ around $x_0$ is included in $\TT$, i.e.
\beq
B_\varepsilon [x_0] = \times_{i=1}^{n/2} [x_{0i}-\varepsilon,x_{0i}+\varepsilon] \subset \TT.
\lb{BCube}
\eeq
Recall that the Gibbs distribution $\P$ has a density \eqref{PCoord}, with 
\beq
P_i(x_i) = \frac{e^{\lambda_{2i-1}x_i + \lambda_{2i} x_i^2}}{\int_0^1 e^{\lambda_{2i-1}t + \lambda_{2i} t^2}\D t},
\lb{Pxi2FeatCoord}
\eeq
for $i=1,\ldots,n/2$. From this and \eqref{BCube} we deduce that 
$$
\P(\TT) \ge \P(B_\varepsilon [x_0]) = \prod_{i=1}^{n/2}
\frac{\int_{x_{0i}-\varepsilon}^{x_{0i}+\varepsilon} e^{\lambda_{2i-1}t + \lambda_{2i} t^2}\D t}{\int_0^1 e^{\lambda_{2i-1}t + \lambda_{2i} t^2}\D t} \to 1,
$$
where the last limit holds if the components of $\bla$ are chosen pairwise, for each feature $i=1,\ldots,n/2$, so that 
$$
\begin{array}{rcl}
\lambda_{2i-i} &\to & \infty,\\
\lambda_{2i} &\to & -\infty,\\
\lambda_{2i-1} + 2x_{0i}\lambda_{2i} &=& 0.
\end{array}
$$
The last displayed equation implies that agent $\A$'s posterior density 
$$
P_i(x_i) = \frac{e^{\lambda_{2i} (x_i-x_{0i})^2}}{\int_0^1 e^{\lambda_{2i} (t-x_{0i})^2}\D t}
$$ 
for coordinate $x_i$ is maximized at $x_{0i}$ and converges weakly to a point mass at $x_{0i}$. Together with the coordinatewise independence \eqref{PCoord}, this implies that $\P$ convergences weakly to a point mass $\bde_{x_0}$ at $x_0$. Since $x_0$ is an interior point of either $\TT$ or $\TT^c$, this implies that it is possible for $\A$ to come arbitrarily close to full learning and full knowledge acquisition.   
\end{proof}

\begin{theorem}\lb{Th:RT}
    Let $\X=[0,1]^r$ and $\PP = \{\AA_1,\ldots,\AA_n\}$ be a finite partition of $\X$ that is obtained as a recursively partitioned binary tree, so that all $\AA_i$ are rectangles with sides parallel to the coordinate axes. Then, full knowledge is only attained if the number of features $n$ goes to infinity.
\end{theorem}

\begin{proof}

To $\X=[0,1]^r$ we assign a uniform prior density $P_0(x)\equiv 1$. The finite partition $\PP = \{\AA_1,\ldots,\AA_n\}$ of $\X$ corresponds to $n$ feature indicator functions $f_i(x) = \1_{\AA_i}(x)$, and the posterior density 
\beq
		P(x) = \sum_{i=1}^n p_i \1_{\AA_i}(x)
	\lb{PPiecewise}
        \eeq
is constant over each $\AA_i$, with values  
%	\beq
%		p_i = n\hmu_i = \frac{ne^{\lambda_i}}{e^{\lambda_1}+\ldots+e^{\lambda_n}} \propto e^{\lambda_i}
%	\lb{pi}
%        \eeq
%that can also be expressed as
\beq
p_i = \frac{\hmu_i}{|\AA_i|} = \frac{e^{\lambda_i}}{|\AA_1|e^{\lambda_1}+\ldots+|\AA_n|e^{\lambda_n}} \propto e^{\lambda_i}.
\lb{pi2}
\eeq
Here $\hmu_i=\hmu_i(\DD)=\P(\AA_i)$ is agent $\A$'s belief about the value of feature $i$ based on data $\DD$, $p_i$ is the value of $P(x)$ on $\AA_i$, and $|\AA_i|=\nu(A_i)$ is the Lebesgue measure of $\AA_i$. Since the feature functions $f_i$ are linearly dependent, without loss of generality we may choose $\bla$ so that the last proportionality of \eqref{pi2} is an equality. 

In order to construct the posterior distribution from a recursively partitioned binary tree, the sets $\AA_i$ must be $r$-dimensional rectangles with sides parallel to the $r$ coordinate axes. In more detail, we make use of a binary tree
$$
\T = \{t_1,\ldots,t_{2n-1}\} = \T_1 \cup \T_2
$$
with $2n-1$ nodes, of which those in $\T_1=\left\{t_1,\ldots,t_n\right\}$ are leaves, those in $\T_2=\left\{t_{n+1},\ldots,t_{2n-1}\right\}$ are interior nodes, and $t_{2n-1}$ is the root of the tree. In particular, $\AA_i$ and $p_i$ are, respectively, a region and a probability weight associated with leaf node $t_i$, for $i=1,\ldots,n$. Each node $t\in\T$ is represented as a binary sequence 
\beq
    t = \left(m_{t1},\ldots,m_{th_t}\right)
    \lb{tDef}
\eeq
of length $h_t$, where $h_t$ is the height of $t$, i.e.\ the number of edges of the path from the root $t_{2n-1}$ to $t$. Edge number $k$ of this path corresponds to a left turn (right turn) if $m_{tk}=0$ ($m_{tk}=1$). The height of the whole tree is the maximal height
$$
    h = \max\left(h_{t_1},\ldots,h_{t_n}\right)
$$
of all leaf nodes, and the tree is balanced if $h=h_{t_i}$ for all leaf nodes. For each $t\in\T$, we define the parental set 
$$
\mbox{pa}(t) = \left\{\begin{array}{ll}
\{(m_{t1},\ldots,m_{t,h_t-1})\}, & t\ne t_{2n-1},\\
\emptyset, & t=t_{2n-1},
\end{array}\right. 
$$
and the offspring set 
$$
\mbox{off}(t) = \left\{\begin{array}{ll}
\emptyset, & t\in\T_1,\\
\{\mbox{ch}_0(t),\mbox{ch}_1(t)\}, & t\in\T_2,
\end{array}\right.
$$
where the two children of an interior node are defined through $\mbox{ch}_l(t)=(m_{t1},\ldots,m_{th_t},l)$ for $l=0,1$. We also define $t(k)=(m_{t1},\ldots,m_{tk})$ as the ($h_t-k$)-fold parent of $t$ for $k=0,\ldots,h_t-1$, with $t(0)=t_{2n-1}$ and $t(h_t-1)=\mbox{pa}(t)$. The set $\AA_i$ and the probability weight $p_i$ are built recursively along the path that connects the root $t_{2n-1}$ with $t_i\in\T_1$. In order to describe this construction in more detail, we associate with each interior node $t\in\T_2$ a splitting coordinate $j_t\in\{1,\ldots,r\}$, a splitting point $a_t\in (0,1)$ and a splitting probability $q_t\in (0,1)$. When $t\in \T_2$ is branched to have two offspring $\mbox{ch}_0(t)$ and $\mbox{ch}_1(t)$, we let 
$$
B_t = \{x\in\X; \, x_{j_t} \ge a_t\}
$$
be the splitting set associated with the right turn $\mbox{ch}_1(t)$, and its complement $B_t^c$ the set that corresponds to the left turn $\mbox{ch}_0(t)$, where $x_{j_t}$ is the $j_t$-th coordinate of $x \in \X$. Then, for each leaf node $t_i\in\T_1$, put
\beq
    \hmu_i = \prod_{k=1}^{h_{t_i}} \left[q_{t_i(k-1)}^{m_{t_{i}k}} \left(1-q_{t_i(k-1)}\right)^{1-m_{t_ik}}\right],
    \lb{piRT}
\eeq
\beq
    \AA_i = \bigcap_{k=1}^{h_{t_i}} \left[\1\left\{m_{t_i(k-1)}=1\right\}B_{t_i(k-1)} + \1\left\{m_{t_i(k-1)}=0\right\}B_{t_i(k-1)}^c \right],
\lb{AiRT}
\eeq
and 
\beq
|\AA_i| = \prod_{k=1}^{h_{t_i}} \left[\left(1-a_{t_i(k-1)}\right)^{m_{t_{i}k}} a_{t_i(k-1)}^{1-m_{t_ik}}\right].
\lb{AiRTVol}
\eeq
From \eqref{pi2}, \eqref{piRT} and \eqref{AiRTVol}, it follows that, without loss of generality, the parameters $\lambda_i$ of the Gibbs distribution $\P$ can be chosen as 
\beq
\begin{array}{rcl}
\lambda_i &=& \log p_i\\
&=& \sum_{k=1}^{h_{t_i}} \left[m_{t_ik}\log q_{t_i(k-1)} + (1-m_{t_ik})\log \left(1-q_{t_i(k-1)}\right)\right]\\
&-& \sum_{k=1}^{h_{t_i}} \left[m_{t_ik}\log \left(1-a_{t_i(k-1)}\right) + (1-m_{t_ik})\log q_{t_i(k-1)}\right]\\
&=& \sum_{k=1}^{h_{t_i}} \left[m_{t_ik}\log \frac{q_{t_i(k-1)}}{1-a_{t_i(k-1)}} + (1-m_{t_ik})\log \frac{1-q_{t_i(k-1)}}{a_{t_i(k-1)}}\right].
\end{array}
\lb{laiRT}
\eeq
If the feature functions $f_i$ are fixed (that is, if $j_t$ and $a_t$ are fixed for all $t\in\T_1$), then agent $\A$ chooses splitting probabilities $q_t$ for all $t\in\T_1$ in order to compute the feature coefficients \eqref{laiRT} of his posterior. %In order to show that full knowledge acquisition requires that the number of nodes $n\to\infty$, we use the fact that 

Since $\PP$ is a partition of $\X$, 
$$
\max_{1\le i \le n} |\AA_i| \ge \frac{1}{n}.
$$
Moreover, since each $\AA_i$ is a rectangle, its diameter satisfies
$$
\mbox{diam}(\AA_i) = \max \{d(x,y)); \, x,y\in\AA_i\} \ge |\AA_i|^{1/r},
$$
where $d(x,y)=\max_{1\le j \le r} |x_j-y_j|$ is the supremum norm in $[0,1]^r$. From the last two displayed equations, we find that
\beq
2\epsilon = \max_{1\le i \le n} \mbox{diam}(\AA_i) \ge \frac{1}{n^{1/r}} \ge \frac{1}{2^{h/r}}, 
\lb{2eps}
\eeq
where the last inequality follows from $n\le 2^h$, with equality for balanced trees. Since all $\AA_i\in\PP$ are rectangles, and the posterior \eqref{pi2} is constant on each $\AA_i$, we deduce from \eqref{2eps} that $x_0\in\X$ can be chosen so that 
\beq
\P(B_\varepsilon(x_0)) < 1. 
\lb{Pless1}
\eeq
We see from \eqref{Pless1} that $n\to\infty$ is a necessary condition in order to guarantee asymptotic full knowledge of $x_0$, i.e., \ $\P(B_\varepsilon(x_0)) \to 1$ as $n\to\infty$ for each $\varepsilon>0$. 
\end{proof}

\begin{proposition}\lb{CounterBill} \ 
    \begin{enumerate}
    \item Let $\AA = \{x_1,x_2,\ldots\} \subset [0,1]$ be a fixed countable set, and define
        \beq
            \G_\A = \s([0,1]\setminus \AA,x_1,x_2,\ldots)
        \lb{GABill}
        \eeq
        as the $\s$-field generated by the complement of $\A$ and the elements of $\A$ (or equivalently, the collection of sets $\BB$ such that either $\BB$ or $\BB^c$ is a subset of $\AA$). Even though it is not possible to express the posterior as a Gibbs distribution, it is sometimes possible to fully learn and acquire full knowledge about a proposition $p$ with the truth set $\TT$. Full learning is possible if either $p$ is true and $\AA\cap\TT\ne \emptyset$ or if $p$ is false and $\AA\cap\TT^c \ne \emptyset$. Full knowledge can be attained if additionally $p$ is true and $x_0\in\AA\cap\TT$, or if $p$ is false and $x_0\in \AA\cap\TT^c$. 
    \item Let 
        \beq
            \tilde{\G}_\A = \s([0,1]\setminus \tilde{\AA},x_1,x_2,\ldots,x_n)
            \lb{GABill2}
        \eeq
        be constructed from the finite set $\tilde{\AA}=\{x_1,\ldots,x_n\}$. Then, it is possible to approximate the posterior with a Gibbs distribution of $n$ features. Full learning is possible under the same conditions as in Part 1, with $\tilde{\AA}$ in place of $\AA$. KA is possible under the same conditions, to a degree that depends on how well the Gibbs distribution approximates the posterior. 
    \end{enumerate}
\end{proposition}

\begin{proof}
Starting with part 1 of the proof, we first observe that $\G_\A$ in \eqref{GABill} is the collection of sets $\BB$ such that either $\BB$ or $[0,1]\setminus \BB$ is a subset of $\AA$. The difference from Billingsley's example is that the set $\AA$ is now fixed, not an arbitrary countable subset of $[0,1]$. Since $\G_\A$ is generated by a countable collection $\PP = \{\AA_0, \AA_1, \ldots\}$ of sets, with $\AA_0 = [0,1]\setminus \AA$ and $\AA_i = \{x_{i}\}$ for $i\ge 1$ we conclude that the probability measure of agent $\A$ must have a density
\beq
P(x) = p_0 + \sum_{i=1}^\infty p_i \delta_{x_i}(x)
\lb{PBill}
\eeq
for some non-negative numbers $p_i$ satisfying $\sum_{i=0}^\infty p_i = 1$. That is, the belief of $\A$ about $x_0$ is a mixture of ignorance (a uniform density with weight $p_0$) and a belief that is supported on $\AA$.  This is to say that data $\DD$ supply $\A$ with information that $x_0$ either belongs to the set $\AA$ or it can be any other element of $[0,1]$. Consider, without loss of generality, the proposition
$$
p: x_0\mbox{ belongs to the set }[0.5,1].
$$
It follows that $f_p(x)=\1_\TT(x)$, with $\TT=[0.5,1]$. Although $\TT\notin \G_A$ and $f_p$ is not measurable with respect to $\G_\A$, if $p$ is true and $\AA\cap\TT\ne\emptyset$ it is still possible for $\A$ to fully learn $p$ (when $p_0=0$ and $p_i=0$ for all $x_i\notin \TT$ in \eqref{PBill}) and additionally acquire full knowledge about $p$ (if also $x_0=x_i\in\AA\cap\TT$ and $p_i=1$). Analogously, if $p$ is false and $\AA\cap\TT^c\ne\emptyset$, it is possible for $\A$ to learn $p$ fully and additionally acquire full knowledge about $p$, if also $x_0\in\AA\cap\TT^c$. However, since $\P$ is constructed as an infinite sum, it is not possible to express \eqref{PBill} in terms of a Gibbs distribution. This proves the first part of the proposition.

To prove part 2, consider the smaller $\s$-field \eqref{GABill2} constructed from the finite set $\tilde{\AA}=\{x_1,\ldots,x_n\}$. This corresponds to a scenario where $\tilde{\G}_\A$ is generated from a finite collection $\PP = \{\AA_0, \AA_1, \ldots,\AA_{n}\}$ of sets, with $\AA_0 = [0,1]\setminus \tilde{\AA}$ and $\AA_i = \{x_{i}\}$ for $1\le i\le n$. It follows that the posterior belief of $\A$ must have a density 
\beq
P(x) = p_0 + \sum_{i=1}^n p_i \delta_{x_i}(x), 
\lb{PBilln}
\eeq
for some non-negative numbers $p_i$ such that $\sum_{i=0}^n p_i=1$. The distribution in \eqref{PBilln} can be approximated by a Gibbs distribution \eqref{Gibbs2} with $n$ features, as follows: Assume $0<x_i<1$ for $i=1,\ldots,n$ and choose $\delta>0$ so small that all $\AA_i(\delta) = [x_i - \delta/2, x_i + \delta/2]$ are disjoint. Then introduce the spiky feature functions
\beq
	f_i(x) = f_i(x;\delta) = \1_{\AA_i(\delta)}(x) \log \delta^{-1}  
\lb{fiBill}
\eeq
for $i=1,\ldots,n$. Let also $\CC (\delta) = [0,1] \setminus \cup_{i=1}^n \AA_i(\delta)$. It follows from \eqref{Gibbs2} that the Gibbs distribution based on features \eqref{fiBill} has a density
\begin{align}
	\begin{aligned}
		P(x) &= Z_{\bla}^{-1} \left[\1_{\CC (\delta)}(x) + \delta^{-1} \sum_{i=1}^n \1_{\AA_i(\delta)}(x) e^{\lambda_i}\right] \\
			&= p_0(\delta) \1_{\CC (\delta)}(x) + \delta^{-1} \sum_{i=1}^n p_i(\delta) \1_{\AA_i(\delta)}(x)\\
			&\stackrel{\mathcal L}{\to} p_0 + \sum_{i=1}^n p_i \delta_{x_i}(x),
\end{aligned}
\label{PxConv}
\end{align}
where $p_0(\delta) = 1 / Z_{\bla}$, $p_i(\delta) = e^{\lambda_i} / Z_{\bla}$ for $i=1,\ldots,n$, and $Z_{\bla} = 1 - n\delta + \sum_{i=1}^n e^{\lambda_i}$. The last step of \eqref{PxConv} refers to weak convergence as $\delta\to 0$, with  
\begin{align}
	\begin{aligned}
		p_0 &= \lim_{\delta\to 0} p_0(\delta) = \frac{1}{1 + \sum_{j=1}^n e^{\lambda_j}}, \\
		p_i &= \lim_{\delta\to 0} p_i(\delta) = \frac{e^{\lambda_i}}{1 + \sum_{j=1}^n e^{\lambda_j}}, \quad i=1,\ldots,n.
	\end{aligned}
\end{align}
\end{proof}

\section{Proofs of results from Section 5 of \texorpdfstring{\cite{DiazEtAl2024}}{Diaz et al. (2024)}}

\begin{proposition}\lb{Prop:SecLearning}
Suppose agent $\tA$ forms his beliefs about agent $\A$'s beliefs in $x_0$ according to the plug-in posterior distribution $\tilde{\P}$, with density
\beq
\tilde{P}(x) = P(x;\hbla) = Q_{\hbla}(x),
\lb{tP}
\eeq
where $\hbla$ is the maximum likelihood estimator of $\bla$, defined in (43) of \cite{DiazEtAl2024}, based on a secondary learning data set $\tilde{\DD}=(x_1,\ldots,x_m)$ of size $m$, an observation of a random sample $\tilde{D}=(X_1,\ldots,X_m)$ with independent components drawn from $\A$'s posterior distribution $\P=\QQ_{\bla}$ in \eqref{Gibbs2}, where $\bla=\bla(\hbmu(\DD))$ is a function of $\A$'s primary data $\DD$. Then asymptotically, $\tA$'s expected learning about agent $\A$'s beliefs in proposition $p$ is    
\beq
\E [\hat{I}^+(\TT)] = I^+(\TT) + \frac{C}{m} + o \left(m^{-1}\right)
\lb{EI+tp}
\eeq
as $m\to\infty$, where expectation is taken with respect to random variations in $\tilde{D}$, whereas $\TT$ is the set of worlds for which $p$ is true. Moreover, $C = \mbox{tr}\left(\J^{-1}\H\right)/2$, $\J = \J(\bla) = \E_{\QQ_{\bla}} \left[\f(X)\f(X)^T\right]$ is the Fisher information matrix that corresponds to the maximum likelihood estimate $\hbla$ of $\bla$, whereas $\H$ is the Hessian matrix of the function $\bla^\prime \to \mbox{Bias}\left(\TT;\bla,\bla^\prime\right)$ at $\bla^\prime=\bla$, with $\mbox{Bias}\left(\TT;\bla,\bla^\prime\right)$ defined in Section 3.2 of \cite{DiazEtAl2024}. Finally, $o\left(m^{-1}\right)$ is a remainder term that is small in comparison to $m^{-1}$ as $m\to\infty$.
\end{proposition}

\begin{proof}[Proof of Proposition \ref{Prop:SecLearning}]
Recall from Section 5.2 of \cite{DiazEtAl2024} that
\beq
    \hat{I}^+(\TT) = I^+(\TT) + \mbox{Bias} ( \TT;\bla,\hbla).
    \lb{I+tP}
\eeq
From the asymptotic theory of maximum likelihood estimates, we find that the estimate $\hbla$ of $\bla$ is asymptotically normally distributed
\beq
    \sqrt{m} (\hbla-\bla) \stackrel{\mathcal L}{\to} N\left(0,\J^{-1}\right)
    \lb{hlaAs}
\eeq
as $m\to\infty$. Insert the normal approximation \eqref{hlaAs} of $\hbla$ into \eqref{I+tP} and perform a second order Taylor expansion of the function $\hbla\to\mbox{Bias}(\TT;\bla,\hbla)$ around $\bla$. After taking expectation of this Taylor expansion, with respect to the normally distributed random variations, we finally obtain \eqref{EI+tp}. 
\end{proof}

\bibliographystyle{imsart-number} % Style BST file (imsart-number.bst or imsart-nameyear.bst)
\bibliography{daangapaBibliography.bib}       % Bibliography file (usually '*.bib')